\documentclass[conference, a4paper]{IEEEtran}
\usepackage{cite}

\ifCLASSINFOpdf
\else
\fi
\usepackage[cmex10]{amsmath}
\usepackage{algorithmic}

\usepackage{amsmath}
\usepackage{amssymb}
\usepackage{amsthm}
\usepackage{mathabx}
\usepackage{textcomp}
\usepackage{url}
\usepackage{upquote}
\usepackage{float}
\usepackage{graphicx}
\usepackage{subcaption}

\usepackage{array}

\usepackage{dblfloatfix}

\usepackage[utf8]{inputenc}
\usepackage{setspace}

\begin{document}
%
\title{Effective Object Tracking in Unstructured Crowd Scenes}
\author{\IEEEauthorblockN{Ishan Jindal}
\IEEEauthorblockA{Electrical Engineering\\
Indian Institute of Technology
Gandhinagar\\
Email: ijindal@iitgn.ac.in}
\and
\and
\IEEEauthorblockN{Shanmuganathan Raman}
\IEEEauthorblockA{Electrical Engineering\\
Indian Institute of Technology
Gandhinagar\\
Email: shanmuga@iitgn.ac.in}}

%
\maketitle
\begin{abstract}
In this paper, we are presenting a rotation variant Oriented Texture Curve (OTC) descriptor based mean shift algorithm for tracking an object in an unstructured crowd scene. The proposed algorithm works by first obtaining the OTC features for a manually selected object target, then a visual vocabulary is created by using all the OTC features of the target. The target histogram is obtained using codebook encoding method which is then used in mean shift framework to perform similarity search. Results are obtained on different videos of challenging scenes and the comparison of the proposed approach with several state-of-the-art approaches are provided. The analysis shows the advantages and limitations of the proposed approach for tracking an object in unstructured crowd scenes.
\end{abstract}


%
\IEEEpeerreviewmaketitle

\section{Introduction}
\label{intro}
\begin{figure*}[htb]
\centerline{%
\begin{tabular}{c@{\hspace{1pc}}c@{\hspace{1pc}}c@{\hspace{1pc}}c}
\includegraphics[width=1.25in]{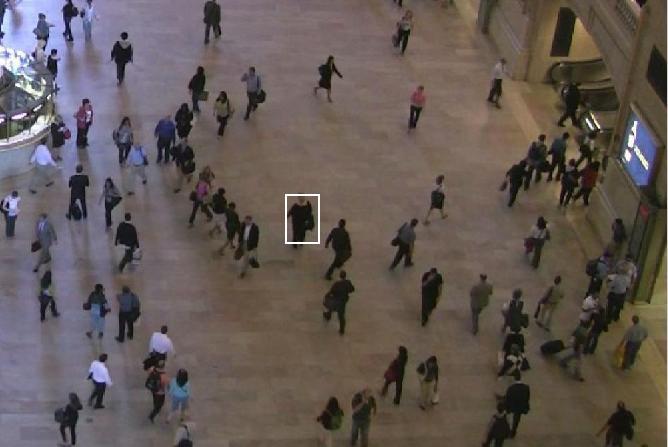} &
\includegraphics[width=1.25in]{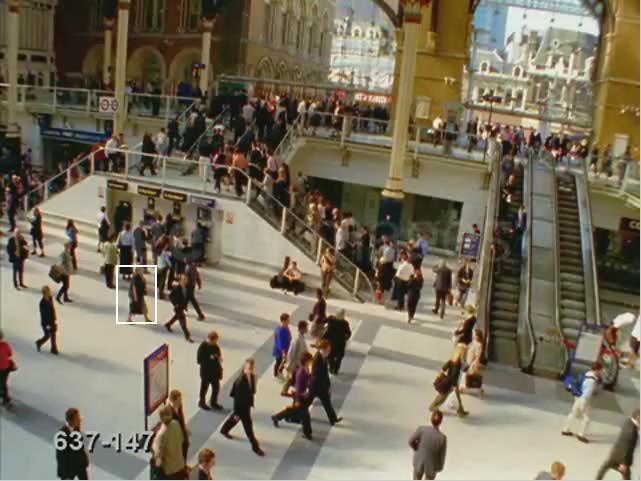}&
\includegraphics[width=1.25in]{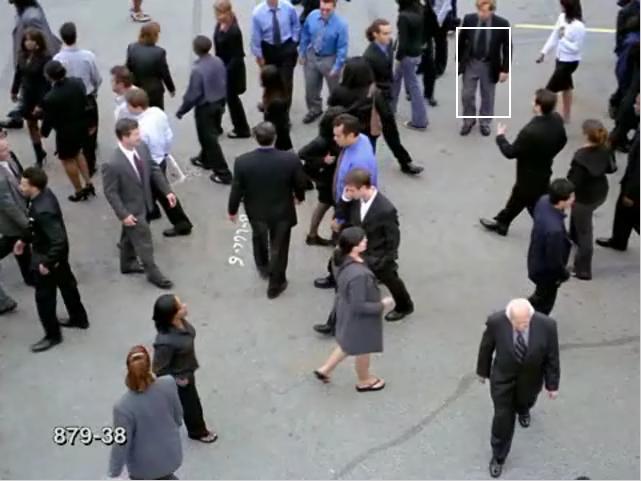} &
\includegraphics[width=1.25in]{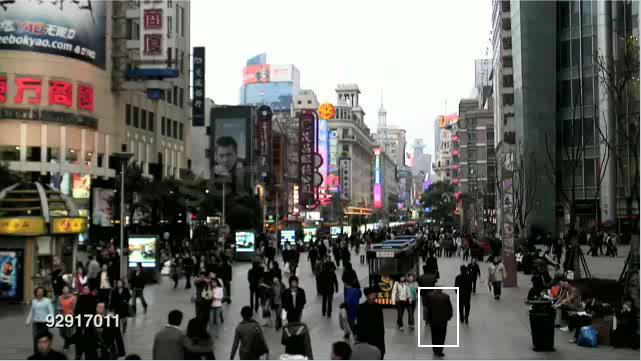} \\ \\
(a)~~UC1&(b)~~UC2&(c)~~UC3&(d)~~UC4
\end{tabular}}
\caption{Test sequences used for evaluation with a white box around the target}
\label{target}
\end{figure*}

Multimedia applications like automatic video surveillance and human computer interaction are among the growing areas of research and tracking of a particular object is an important problem. Most of the available tracking algorithms perform poorly when it comes to tracking in unstructured crowd regions as shown in Fig.\ref{target}. This problem becomes more severe with the increase in crowd density and a significant degradation of the algorithm's performance can be noticed. Increase in crowd density (marathon, crosswalk, rallies) lead to structured motion \cite{floorfildes} where all the objects follow a similar pattern and follow a fixed direction of motion. Whereas, the medium density scenes (shopping mall, exhibition center, railway station, airport) lead to unstructured crowd motion where the direction of motion appears to be random for every spatial location. As these unstructured crowd scenes are more common than the structured ones so tracking in unstructured crowd scenes becomes a very important task. It is relatively very difficult for any tracking algorithm to track a particular person in a crowd especially in medium crowd scenes rather than tracking in highly dense scenes which are more structured \cite{rodriguez2011data}. 

Also, most of the available tracking methods use color distribution of the target for tracking. The color distribution is very volatile and can vary over time because of the changes in illumination and camera parameters. So, using color distribution for tracking in a unstructured crowd scene is not a good choice because it can be possible that the target may encounter or interact with another object having the same color.

In this paper, our main focus is to develop an algorithm that can handle unstructured crowd scenes effectively and efficiently. The main contribution of our work are listed below.
\begin{enumerate}
\item An oriented texture curve (OTC) feature descriptor systematically incorporated with the mean shift estimator for tracking  \cite{margoinOTC}.

\item The performance of the resulting system is discussed in detail through results. This work describes an effective approach for tracking and demonstrates the use of mean shift vector in tracking under unstructured crowd scenes.
\end{enumerate}

This paper is organized as follows. Section 2 deals with the literature review, which describes the number of methods and their merits and demerits with respect to the tracking in unstructured crowd scenes. Section 3 describes the overall approach we adopt towards solving this challenging problem. Section 4 illustrates the experiments done using different datasets for the experiments. It also discusses the results obtained using the proposed approach and compares with the several state-of-the-art approaches. We conclude the paper in section 5 listing some of the improvements which can be include to the proposed approach in future. 
\label{sect:1}
\section{Related Work}
Tracking objects especially humans is an important component of computer vision applications like motion based segmentation, automated surveillance, human computer interaction and traffic monitoring. Various attempts have already been made in order to perfectly detect and track the objects. All the methods developed are different from each other in their approach towards the problem. According to  \cite{yilmaz2006object}, it include point to point tracking, kernel based tracking and silhouette tracking.

Point to point tracking or feature tracking performs tracking in long sequences based upon the number of salient features. These methods first detect the features of target and then search for similar features in the subsequent frames. For these methods, it is required to capture the features that uniquely describe the target and these methods generally fail in the presence of multiple similarities. Classical approaches for feature based tracking include particle filter \cite{particalT}, Kalman filtering \cite{kalmanBroida}, hidden Markov models and  adpative color based particle filter \cite{nummiaro2003adaptive}, to name a few.

Kalman filter tracking comprises of two steps: prediction and correlation. The first step is used to predict the new states of the variables and the next step updates the state of the object based on the current observation. In the case of non-linearity and when the state variables do not follow Gaussian distribution, the performance of particle filter is more robust than the Kalman filter. An adaptive color based particle filter follows a top down approach and gives good results because of its sensitivity towards the color of target \cite{nummiaro2003adaptive}. These filters recently became popular in tracking and detection but fails in highly cluttered videos like in the case of unstructured crowd scenes. Also, these methods fail when tracking multiple objects because of the possibility of incorrect correspondences. These problems are tackled by statistical data association techniques such as  multiple hypothesis tracking (MHT) and probabilistic MHT \cite{mltracking}.

Kernel based tracking algorithms are based on the object motion and typically include techniques like mean shift tracker \cite{comaniciu2003kernel}, Lucas-Kanade tracker (KLT) \cite{KLTshi}, Extended mean shift tracker \cite{zivkovic2004like} and Approximate Bayesian method \cite{zivkovic2009approximate}. Both mean shift and extended mean shift are density based object motion model. Extended mean shift is an extension of mean shift tracker which tries to find the position of local maximum of density function. In all the methods \cite{comaniciu2003kernel,KLTshi,zivkovic2004like} the target is represented as a weighted histogram of color values and resulting in same problem as discussed in section \ref{sect:1} . Some other methods like SIFT \cite{objecttrac} and spatio-temporal \cite{cannons2007spatiotemporal} descriptors have been developed to represent the target. Different similarity measures like Bhattacharyya coefficient, Kullback-Leibler (K-L) divergence can be employed to detect the target in subsequent frames. KLT Tracker iteratively computes the translation over the selected region, where the translation is computed by optical flow methods \cite{hornschunk,lukas}. In the proposed approach we have used mean shift tracking because of its accuracy, robustness and stability for tracking coloured objects. Other training based tracking methods have also been proposed like Adaptive background mixture model  \cite{GrimsonStauffer} and support vector machine (SVM) classifier \cite{svt}, which intelligently distinguishes between the images of the objects to be tracked from all the other objects in the scene which should not be tracked. 

For crowd tracking algorithms which include floor fields and correlated topic model, it can be inferred from \cite{floorfildes} that floor fields can only be used in the case of tracking in structured motion where there is one and only one dominant direction of crowd motion. This method fails for unstructured crowd tracking when each location has multiple dominant crowd motion directions. Also, \cite{trackunstr} is an attempt to tackle the particular problem of tracking in unstructured crowd scenes and makes use of correlated texture model (CTM). 

\section{Proposed Approach}
In real time tracking, it is obvious that the target can interact with other persons or can be bypassed by the other person coming from a different direction. It can also be possible that the target is entirely occluded by another person wearing the similar attires and these situation occur very frequently in the case of unstructured crowd scenes such as shopping mall, busy street and exhibition centers. 

These problems can be effectively addressed by the following proposed approach. In the subsequent subsections, we'll describe the  proposed approach starting from the feature descriptor to the mean shift target localization. 
\subsection{Texture based descriptor}

\subsubsection{Oriented texture curves}
Oriented texture curves are low level descriptors (OTC) generally used for scene classification \cite{margoinOTC}. Given an image or a target, patches are sampled along highly dense grid, and the resulting patch is represented by a number of curves most commonly by eight curves corresponding to eight orientations. Each curve represents the property of the patch along a particular direction and each curve is characterized by a novel curve descriptor. Concatenating all the directional curve descriptors results in a single descriptor for a dense grid patch. As shown in the Fig. \ref{target}, target is manually selected and the texture descriptor is computed for the target over a densely grid of size $3\times3$.

Firstly, the target is divided into patches of size $N\times N$ and the obtained patch is further divided into $N$ strips along different orientations. In order to capture the different features, patch is examined in all the orientations as described in \cite{localdis}. For a $N\times N$ patch we obtain eight $N$ point curves corresponding to all eight directions.
\begin{equation}
v_\theta(i) = \frac{1}{|K_{\theta,i}|} \sum_{x \epsilon K_{\theta,i}} P(x) \hspace{2.3cm}   1\leq i \leq N
\end{equation}
Here, $v_\theta$ is the $N$ point curve for a patch P in the orientation $\theta$ and $v_\theta(i)$ denotes the $i^{th}$ point of the curve $v_\theta$. $|K_{\theta,i}|$ is the number of pixels contained within $i^{th}$ strip having $\theta$ orientation. Here, the property of rotational sensitivity is employed using the predefined patch orientations, regardless of the patch itself. The obtained curves are illumination sensitive and not robust to geometric differences. As we are interested only in the shape of the curve, the gradient $v'_\theta(i)$ and the approximation to its curvature $v''_\theta(i)$ are used, which best describes the shape of the curve \cite{curves}.
 \begin{equation}
 v'_\theta(i)=v_\theta(i+1)-v_\theta(i), \hspace{0.2cm} 1\leq i < N
 \end{equation}
\begin{equation}
 v''_\theta(i)=v'_\theta(i+1)-v'_\theta(i), \hspace{0.2cm}  1\leq i < (N-1)
\end{equation}
Similarly for the RGB images, RGB gradients are computed by calculating the $l_2$ distance between them and having same sign as that of their gray scale gradients.
\begin{equation}
 V'_\theta(i)=sign\{v'_\theta\}.||V_\theta(i+1)-V_\theta(i)||_2, \hspace{0.2cm}  1\leq i < N
\end{equation}
\begin{equation}
 V''_\theta(i)=V'_\theta(i+1)-V'_\theta(i),   \hspace{0.2cm}  1\leq i < (N-1)
\end{equation}
where, $V'_\theta(i)$ denote the gradients in the case of an RGB image and $v'_\theta(i)$ denote the gradients in the case of a gray scale image. A descriptor $D$ for a single patch is obtained by concatenating all the gradients and curvatures. For a $13\times13$ patch, length of descriptor is obtained as $8 \times (12+11)= 184$ values for one patch. Resulting descriptor is robust to illumination differences as well as to the geometric distortions. Now, in order to differentiate between the textured and texture-less regions, (i.e. to obtain a descriptor robust to local contrast differences), a homogeneous-bin (H-bin) normalization scheme is employed \cite{histogram}. Here an H-bin (small valued bin = 0.05) is appended to the overall descriptor $D$ of both texture and texture less regions resulting in a total length of $1+184=185$ values per patch and then $l_2$ normalization leads to an effective textured feature descriptor $f$, described by $
f=\frac{\left(H-bin, D\right)}{||\left(H-bin, D\right)||_2}$.

\subsubsection{Target Representation}
For the target representation in each of the videos shown in Fig.\ref{target} OTC are calculated. The obtained local descriptor space is very large so it is required to partition it into the informative regions and all  these regions are combined to form a vocabulary of visual patches using $K$-means clustering \cite{Devil_coding}. For $n$ number of feature descriptors $(f_1,f_2,.....,f_n)  \in R^d $, $k$ is the number of visual words vocabulary $(\mu_1,\mu_2,......,\mu_k)\in R^d $ generated. We try to minimize the cumulative approximation error ($E$), where $d$ is the dimension of the feature descriptor $E = \sum_{i=1}^n ||f_i-\mu_i||^2$.

In order to obtain the compact representation of the target, baseline histogram encoding scheme is employed which is obtained using hard quantization. Fig. 2 shows the codebook encoding of the target of image scene 1 shown in Fig. \ref{target} using vector (hard) quantization (VQ). VQ is done by alternating between seeking the best means which match the given assignments $\mu_k=avg\{f_i: O_i=k\}$. 
Given the best means, we are seeking for best assignments using $O_{ki}= \min_k  \sum_{i=1}^k ||f_i-\mu_k||^2$.

This results in $\hat{O}=O_{ki}$ the probability density function for the object to be tracked. In the similar way probability density function for the candidate target model $\hat{C}$ in the current frame is calculated.  
\begin{figure}[]
\begin{center}
\includegraphics[width = 2.5in]{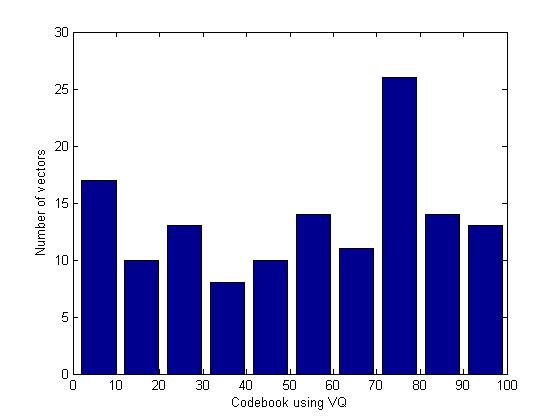}
\caption{Codebook PDF for the target region in Fig.\ref{target}.a}
\end{center}
\label{tarhist}
\end{figure}

\begin{figure*}[htbp]
\centerline{%
\begin{tabular}{c@{\hspace{1pc}}c@{\hspace{1pc}}c@{\hspace{1pc}}c}
Frame 1 & Frame 23 & Frame 43 & Frame 63\\ \\
\includegraphics[width=0.12\linewidth]{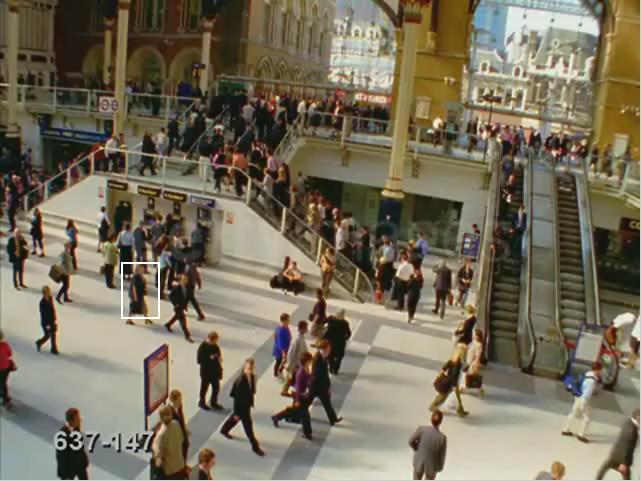} &
\includegraphics[width=0.12\linewidth]{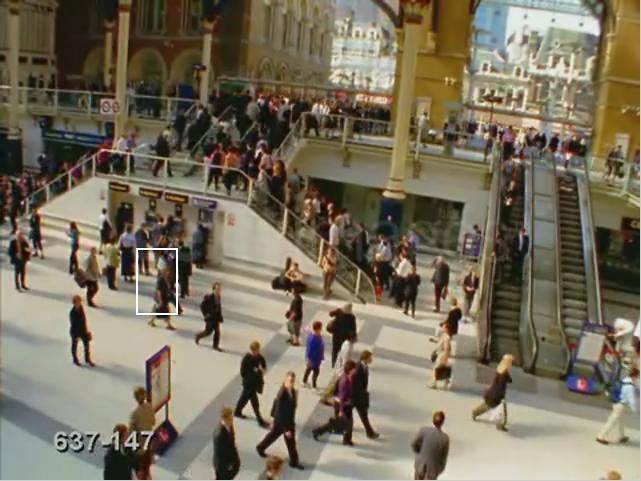} &
\includegraphics[width=0.12\linewidth]{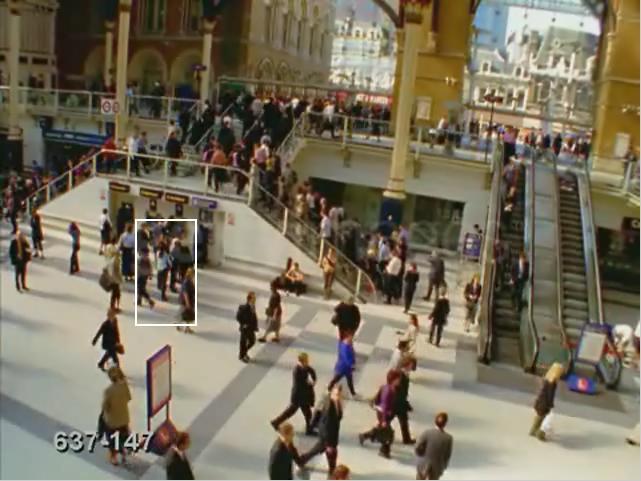} &
\includegraphics[width=0.12\linewidth]{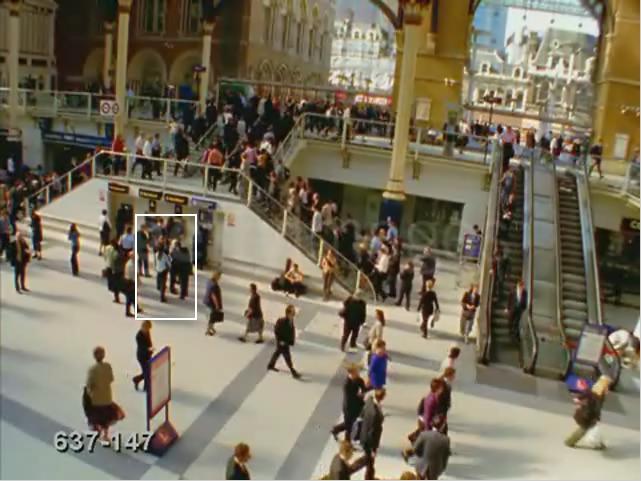} \\
\includegraphics[width=0.12\linewidth]{EMS_1_crowd_2} &
\includegraphics[width=0.12\linewidth]{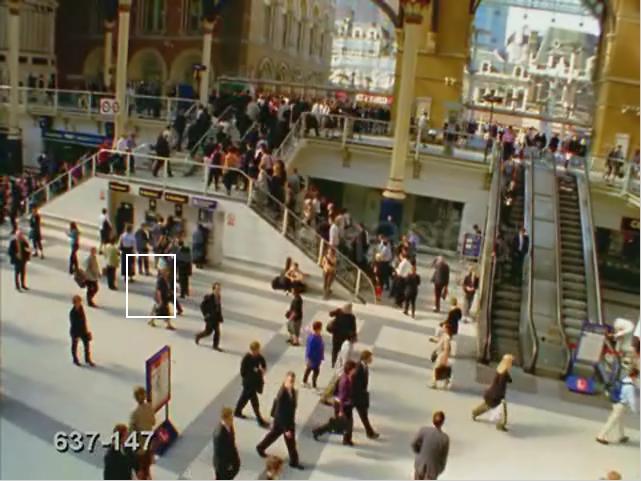} &
\includegraphics[width=0.12\linewidth]{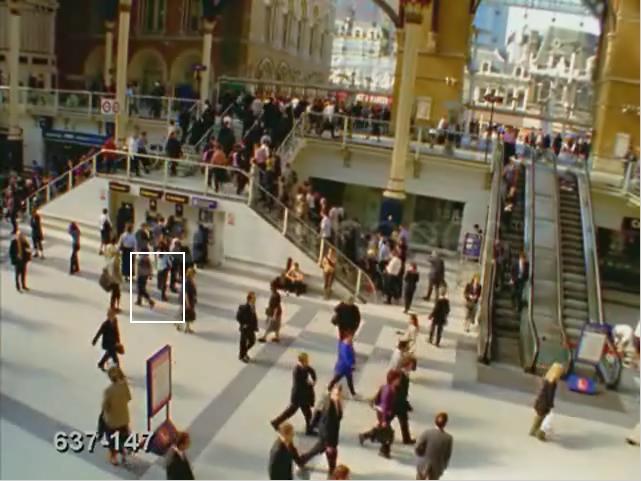} &
\includegraphics[width=0.12\linewidth]{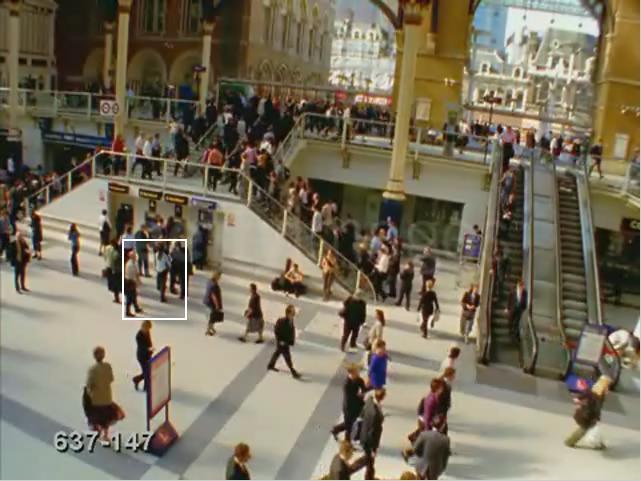} \\
\includegraphics[width=0.12\linewidth]{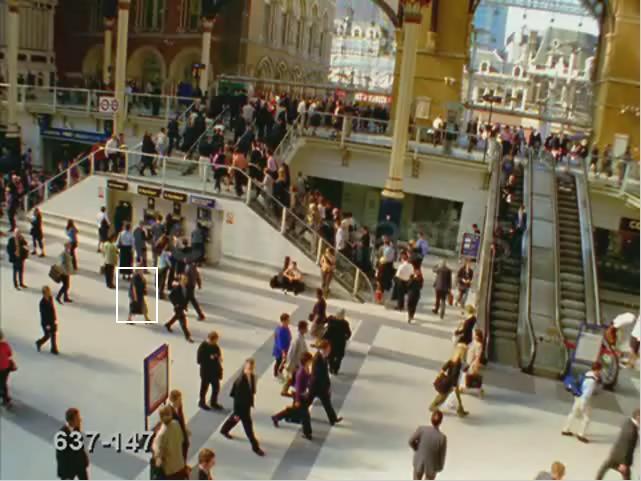} &
\includegraphics[width=0.12\linewidth]{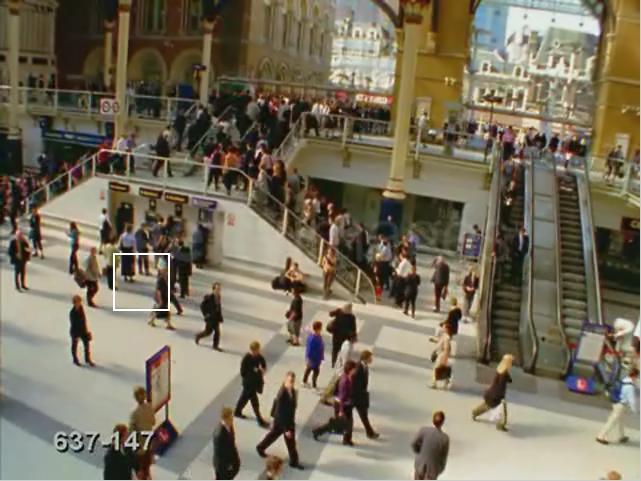} &
\includegraphics[width=0.12\linewidth]{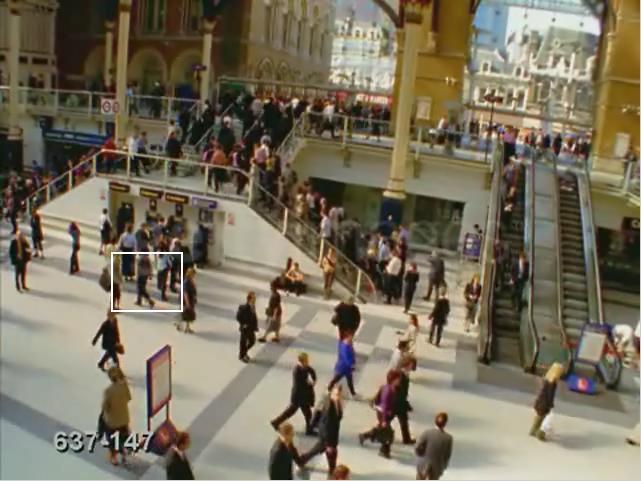} &
\includegraphics[width=0.12\linewidth]{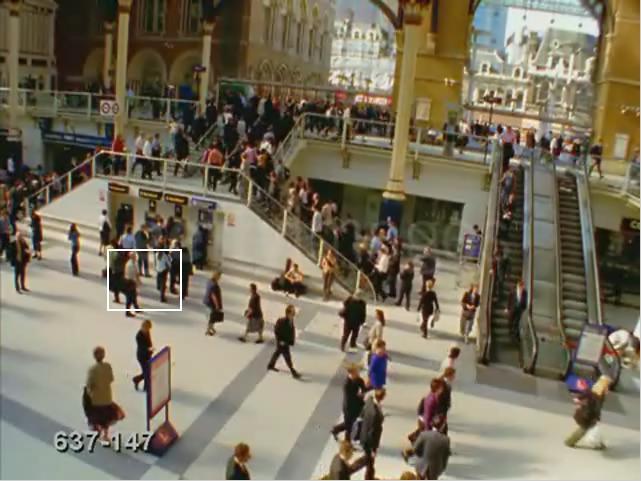} \\
\includegraphics[width=0.12\linewidth]{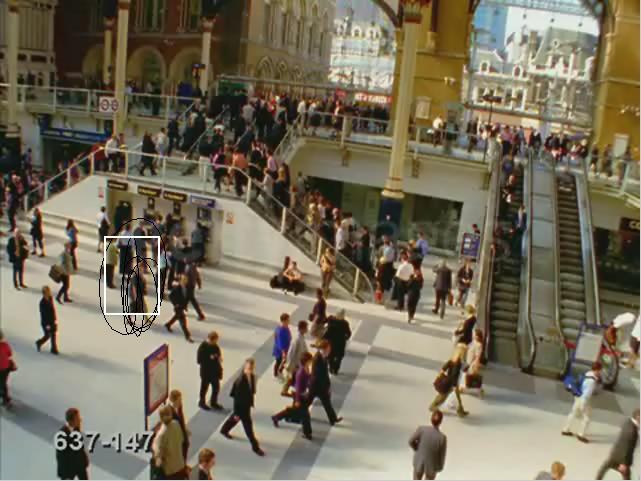} &
\includegraphics[width=0.12\linewidth]{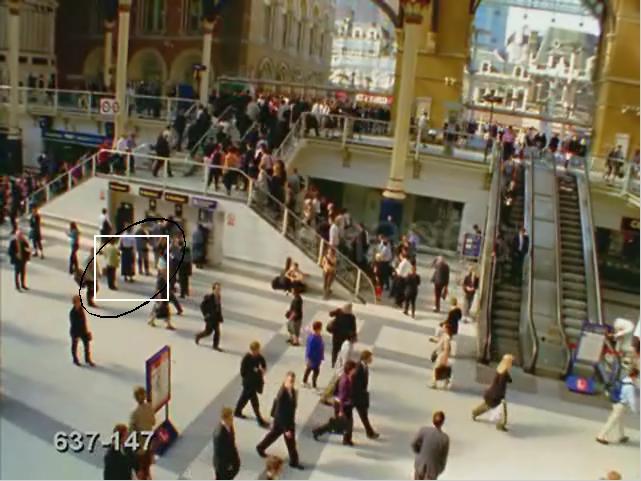} &
\includegraphics[width=0.12\linewidth]{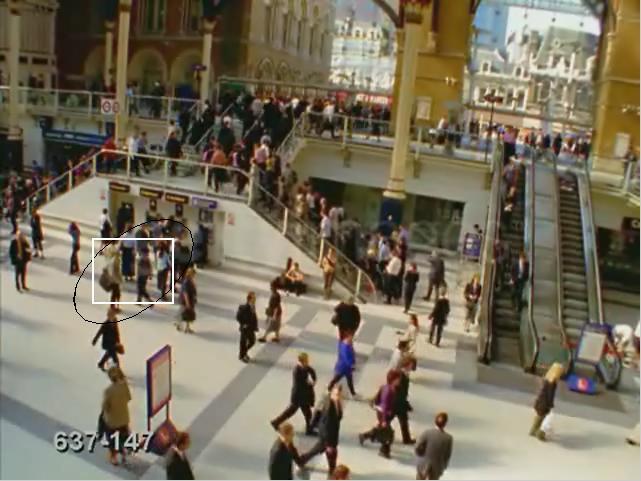} &
\includegraphics[width=0.12\linewidth]{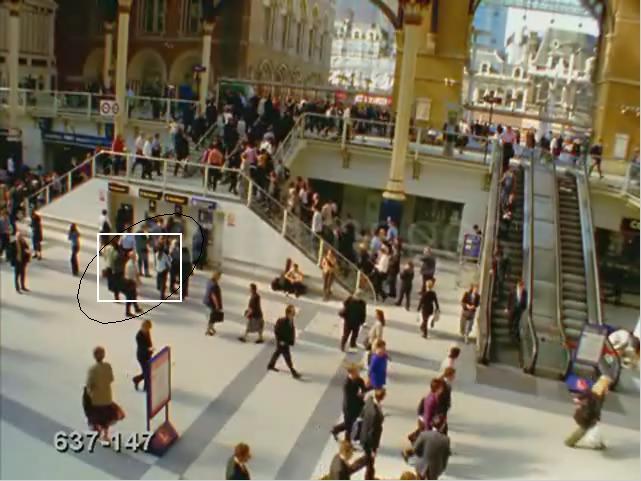} \\
\includegraphics[width=0.12\linewidth]{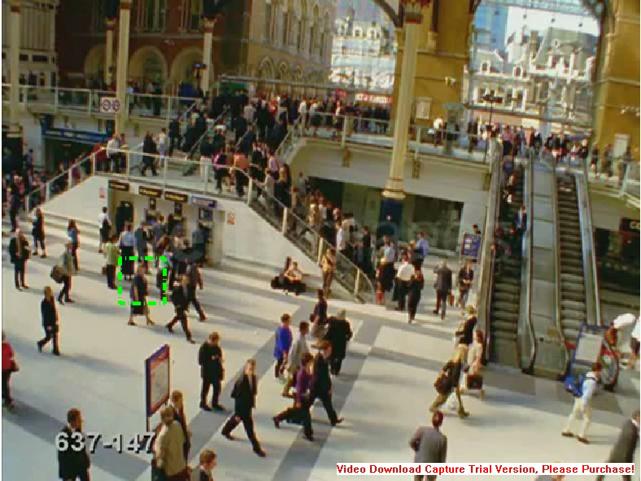} &
\includegraphics[width=0.12\linewidth]{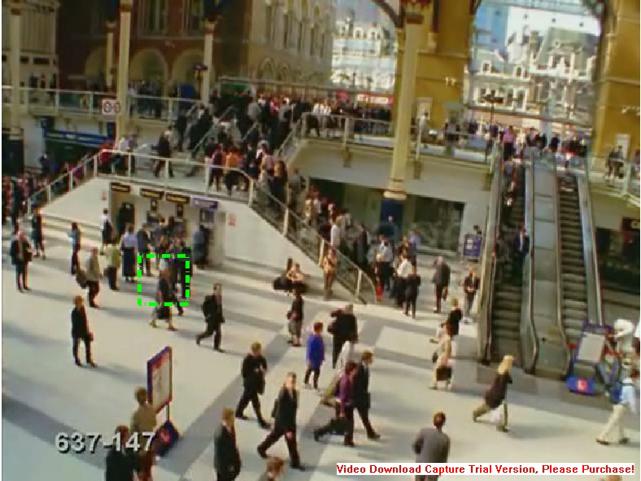} &
\includegraphics[width=0.12\linewidth]{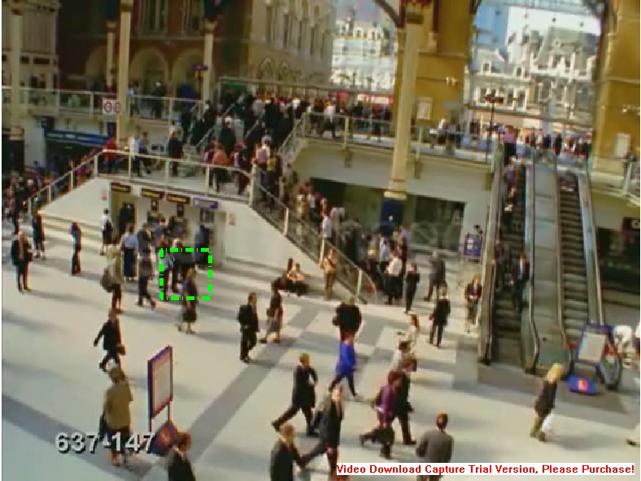} &
\includegraphics[width=0.12\linewidth]{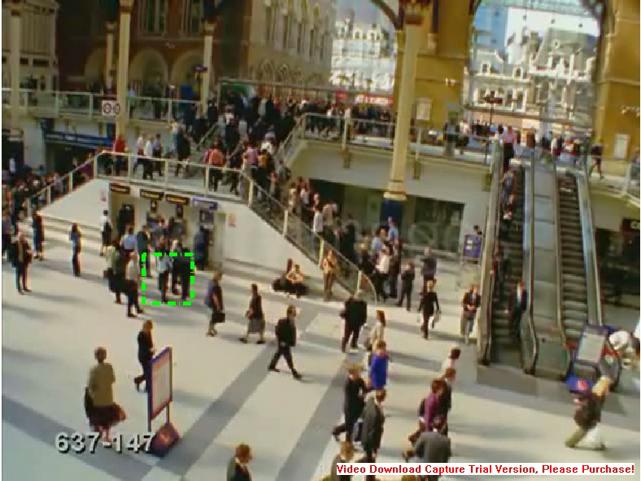} \\
\includegraphics[width=0.12\linewidth]{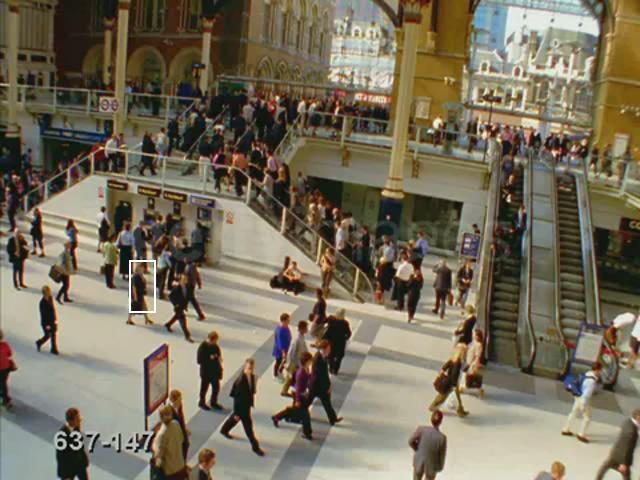} &
\includegraphics[width=0.12\linewidth]{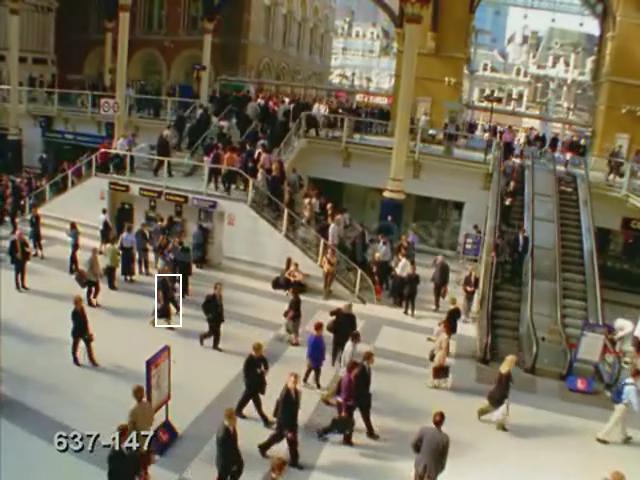} &
\includegraphics[width=0.12\linewidth]{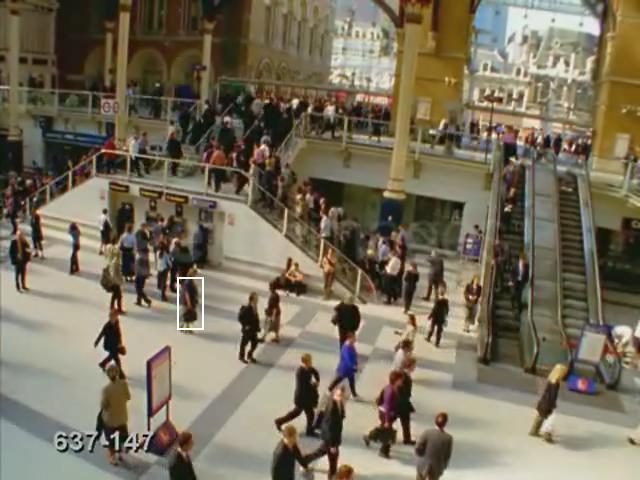} &
\includegraphics[width=0.12\linewidth]{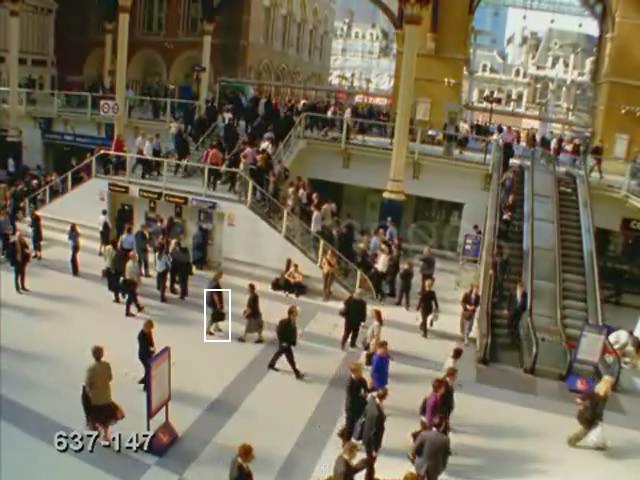} 
\end{tabular}}
\caption{Test sequences 2: Comparison of MS (1st row) \cite{comaniciu2003kernel}, EMS (2nd row) \cite{zivkovic2004like}, APF (3rd row) \cite{nummiaro2003adaptive}, MKF (4th row) \cite{zivkovic2009approximate}, CTM (5th row) \cite{trackunstr} and the proposed approach (Last row).}
\label{sequence2}
\end{figure*}
\begin{figure*}[htbp]
\centerline{%
\begin{tabular}{c@{\hspace{1pc}}c@{\hspace{1pc}}c@{\hspace{1pc}}c}
Frame 23 & Frame 58 & Frame 91 & Frame 139\\ \\
\includegraphics[width=0.12\linewidth]{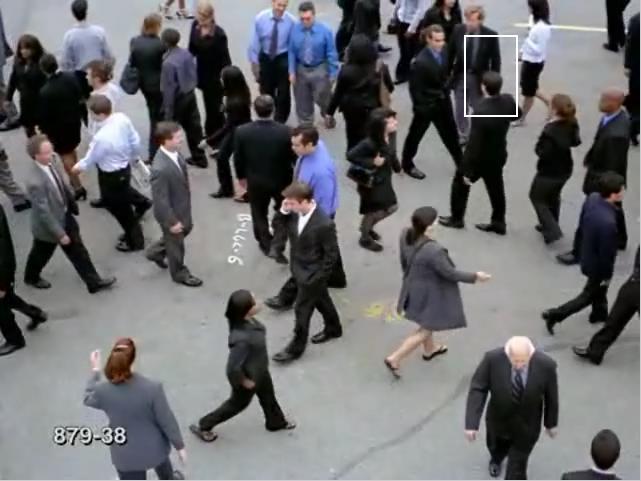} &
\includegraphics[width=0.12\linewidth]{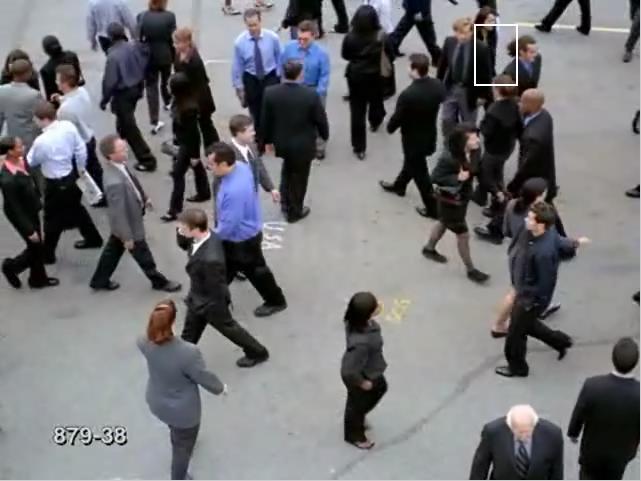} &
\includegraphics[width=0.12\linewidth]{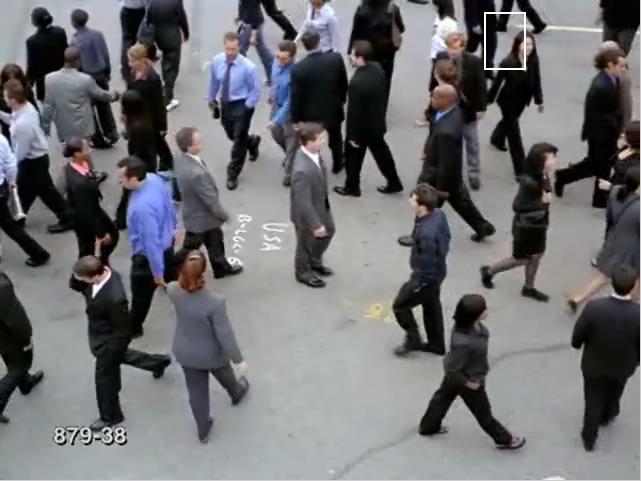} &
\includegraphics[width=0.12\linewidth]{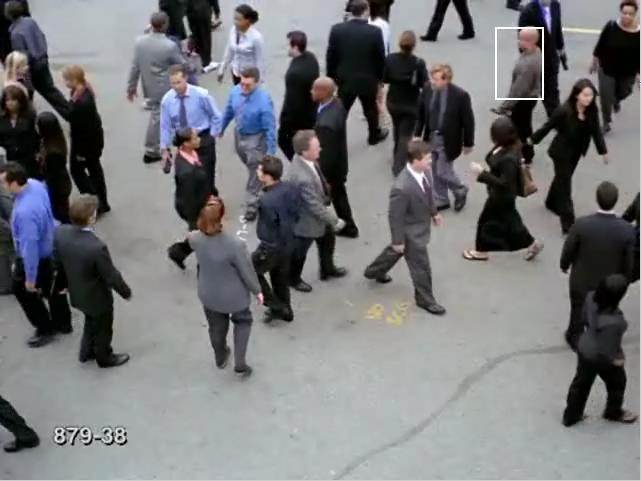} \\
\includegraphics[width=0.12\linewidth]{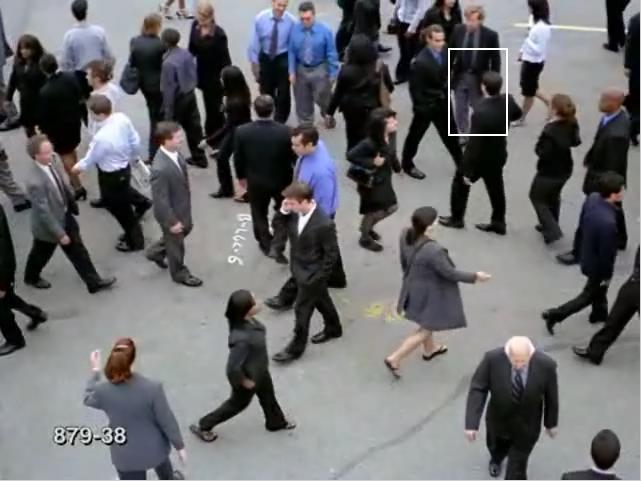} &
\includegraphics[width=0.12\linewidth]{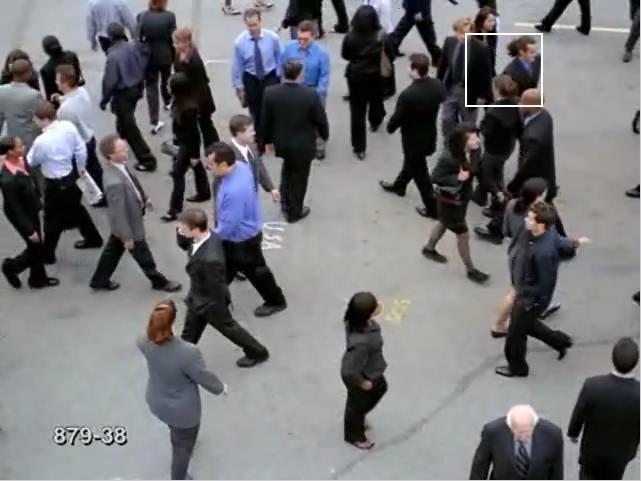} &
\includegraphics[width=0.12\linewidth]{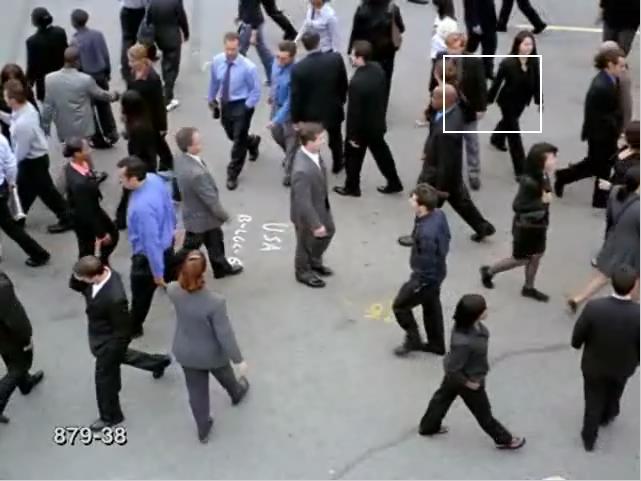} &
\includegraphics[width=0.12\linewidth]{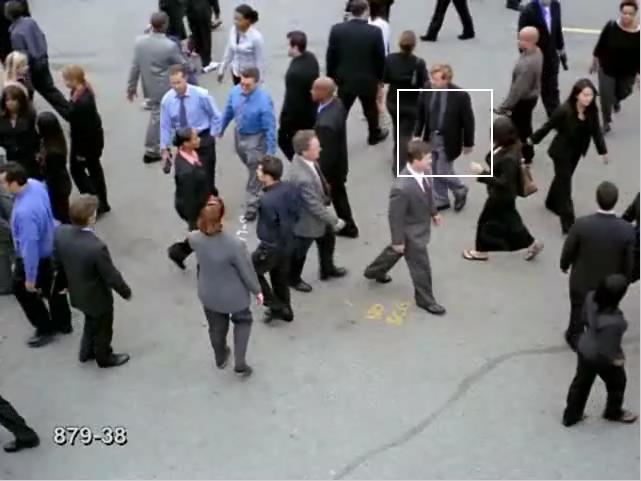} \\
\includegraphics[width=0.12\linewidth]{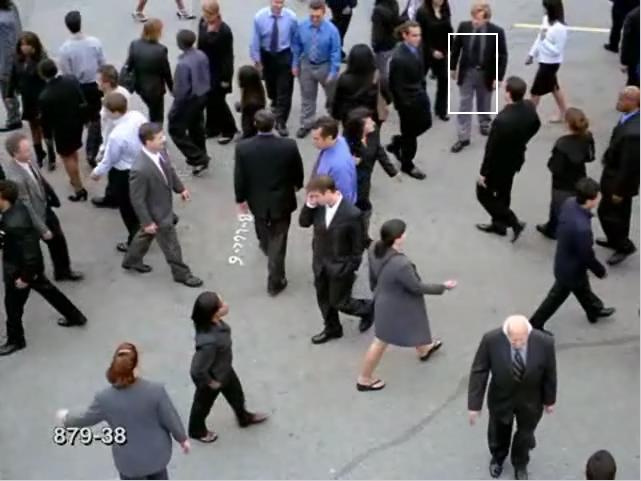} &
\includegraphics[width=0.12\linewidth]{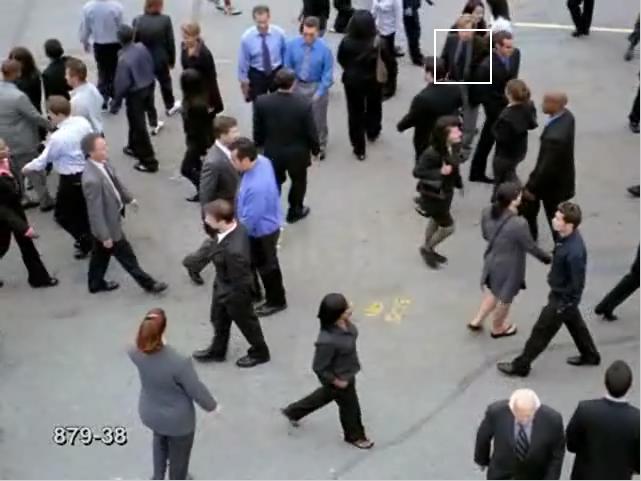} &
\includegraphics[width=0.12\linewidth]{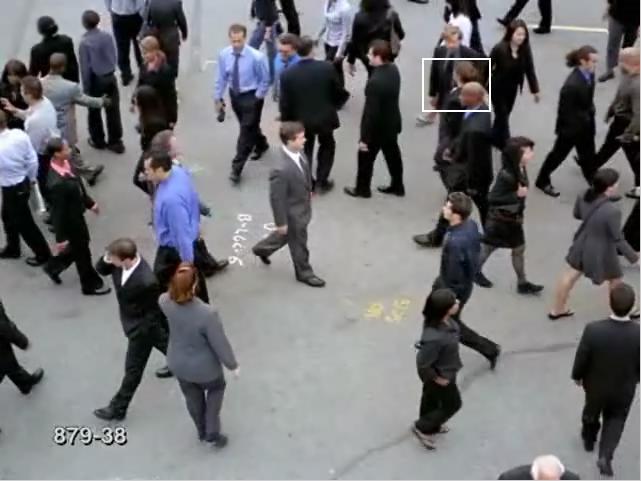} &
\includegraphics[width=0.12\linewidth]{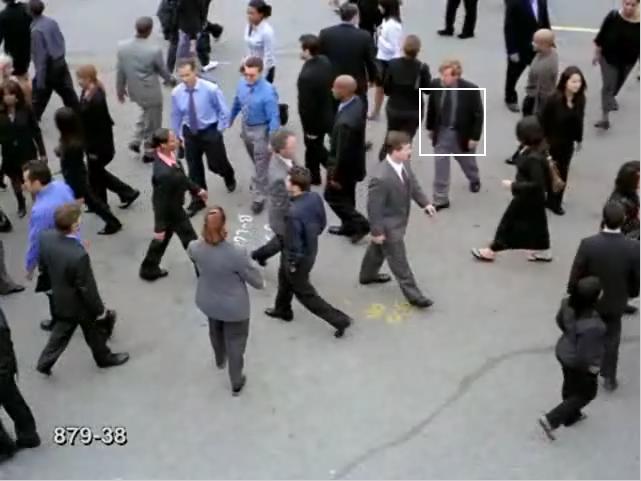} \\
\includegraphics[width=0.12\linewidth]{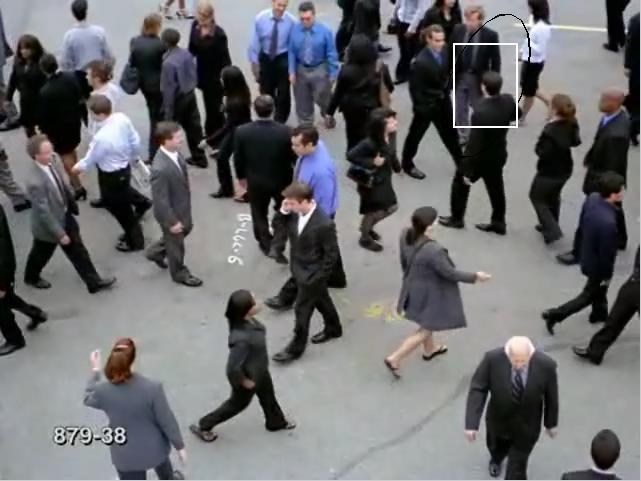} &
\includegraphics[width=0.12\linewidth]{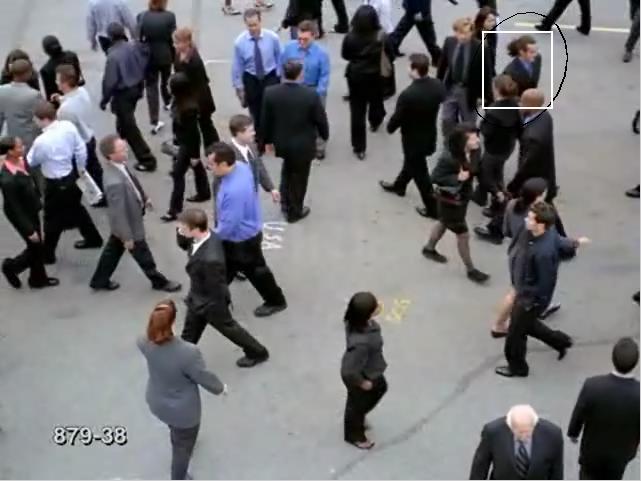} &
\includegraphics[width=0.12\linewidth]{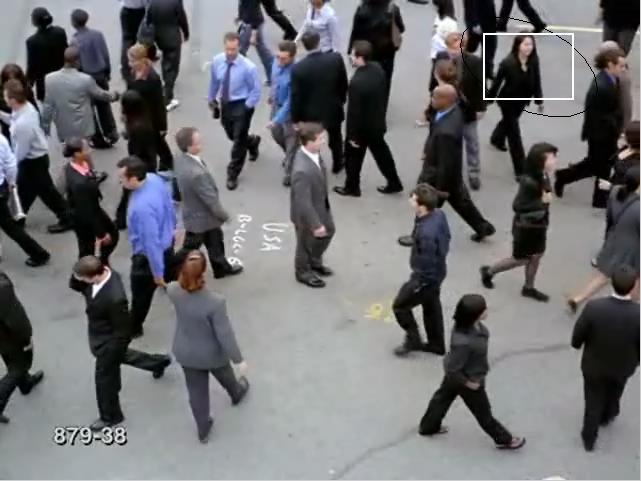} &
\includegraphics[width=0.12\linewidth]{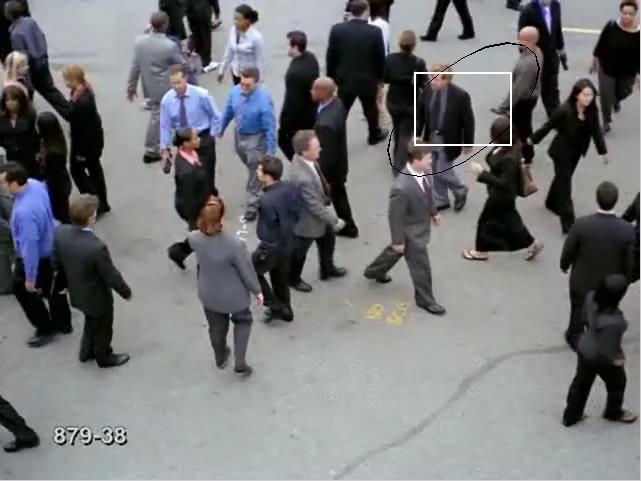} \\
\includegraphics[width=0.12\linewidth]{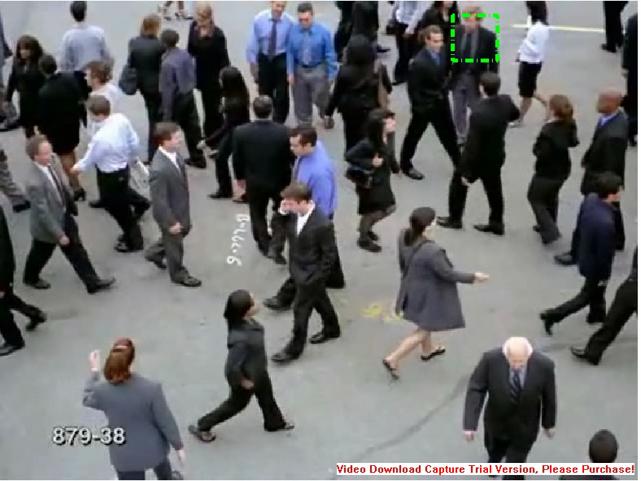} &
\includegraphics[width=0.12\linewidth]{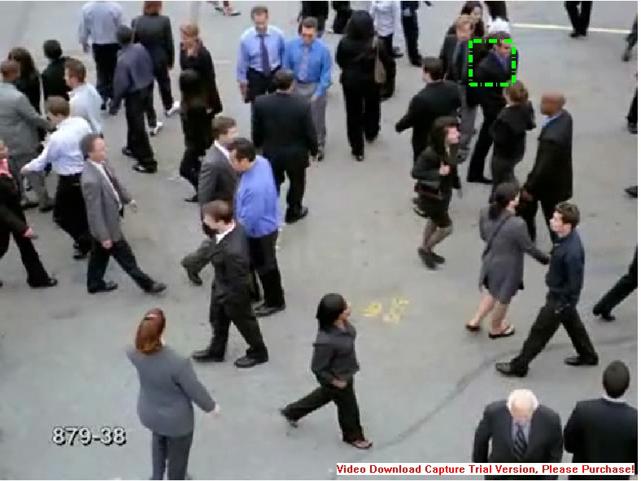} &
\includegraphics[width=0.12\linewidth]{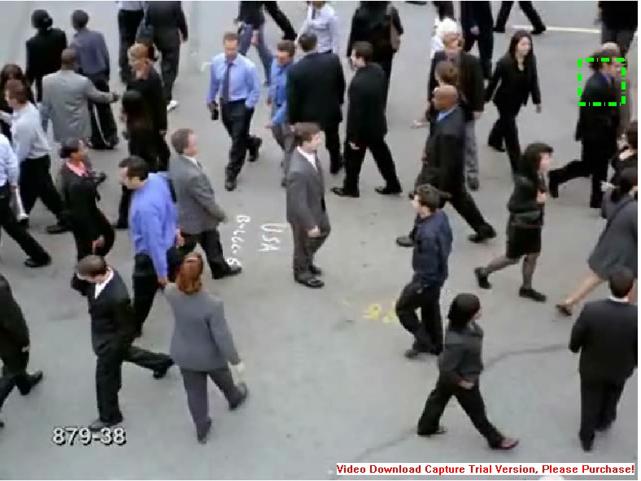} &
\includegraphics[width=0.12\linewidth]{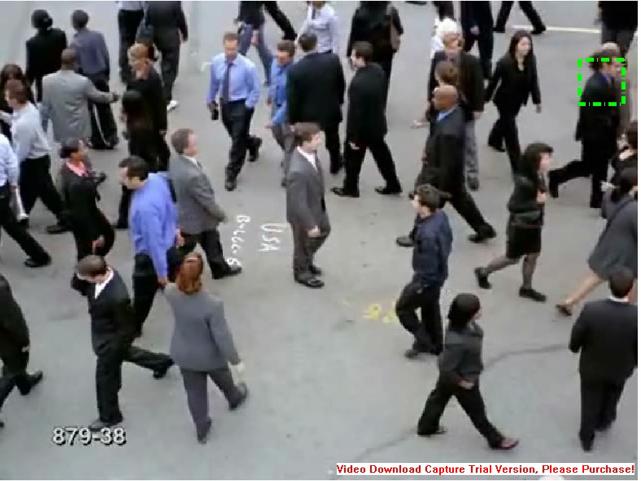} \\
\includegraphics[width=0.12\linewidth]{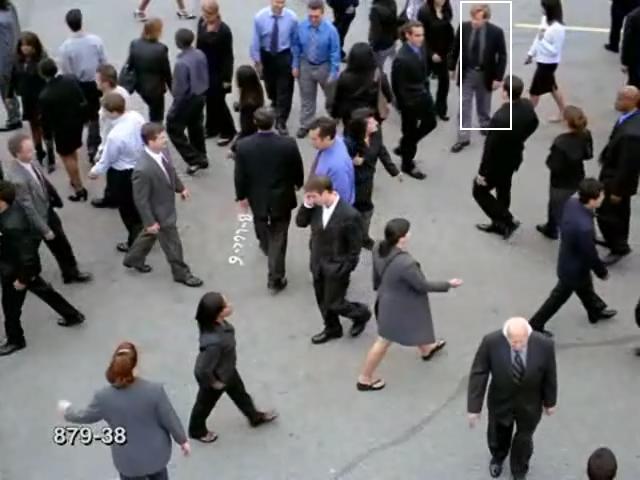} &
\includegraphics[width=0.12\linewidth]{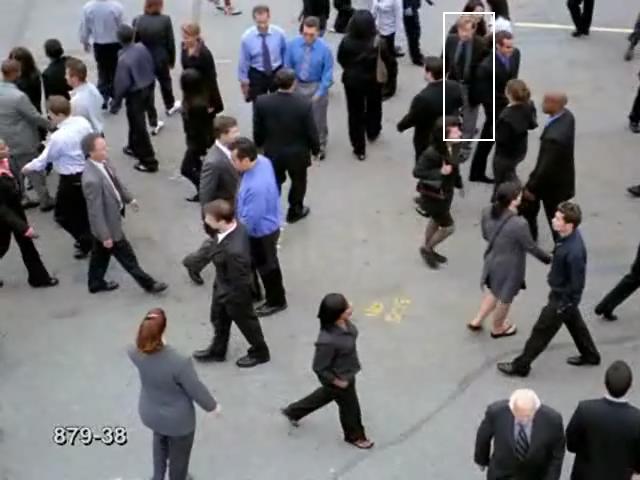} &
\includegraphics[width=0.12\linewidth]{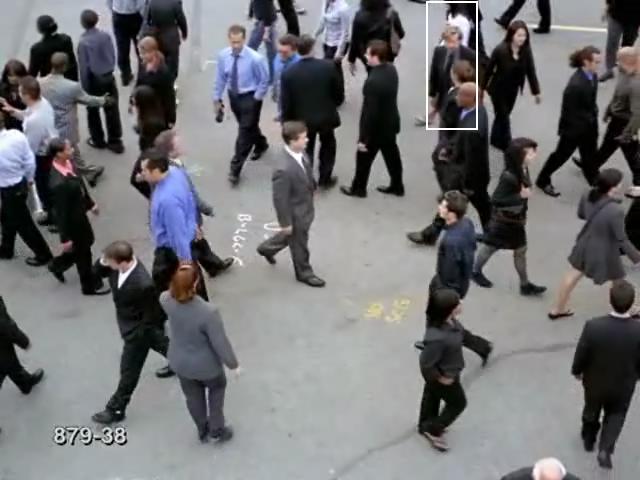} &
\includegraphics[width=0.12\linewidth]{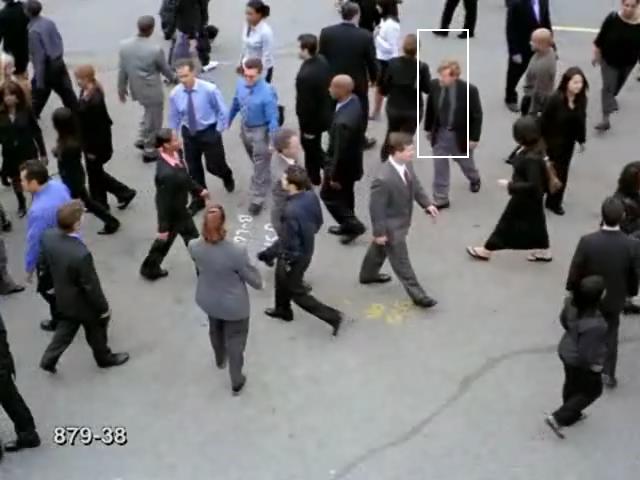} 
\end{tabular}}
\caption{Test sequences 3: Comparison of MS (1st row) \cite{comaniciu2003kernel}, EMS (2nd row) \cite{zivkovic2004like}, APF (3rd row) \cite{nummiaro2003adaptive}, MKF (4th row) \cite{zivkovic2009approximate}, CTM (5th row) \cite{trackunstr} and the proposed approach (Last row).}
\label{sequence3}
\end{figure*}
\begin{figure*}[htbp]
\centerline{%
\begin{tabular}{c@{\hspace{1pc}}c@{\hspace{1pc}}c@{\hspace{1pc}}c}
Frame 42 & Frame 82 & Frame 112 & Frame 132\\ \\
\includegraphics[width=0.12\linewidth]{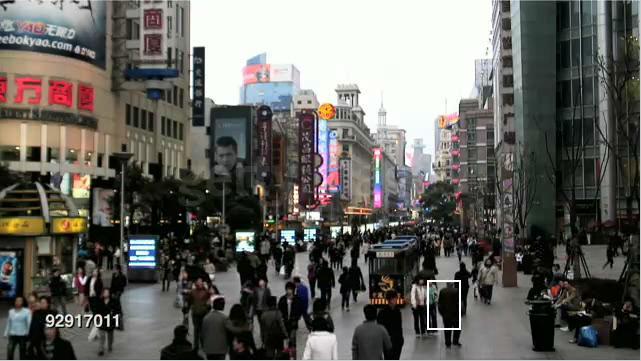} &
\includegraphics[width=0.12\linewidth]{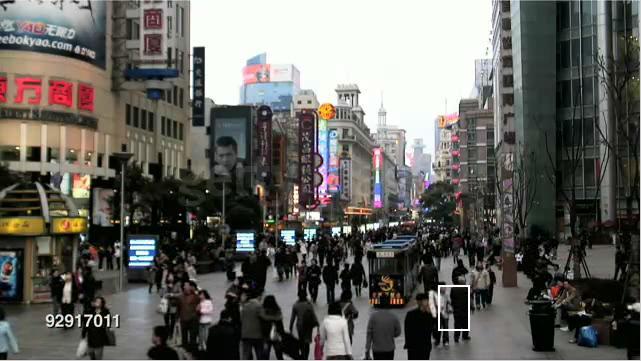} &
\includegraphics[width=0.12\linewidth]{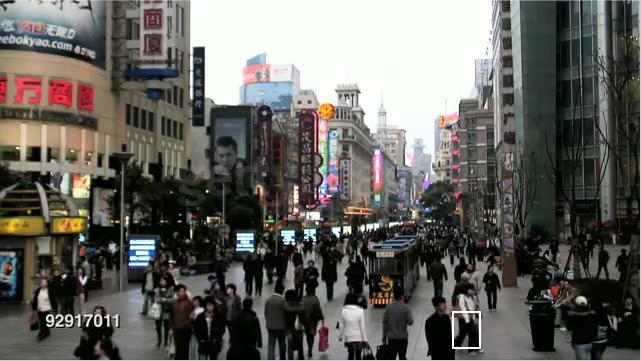} &
\includegraphics[width=0.12\linewidth]{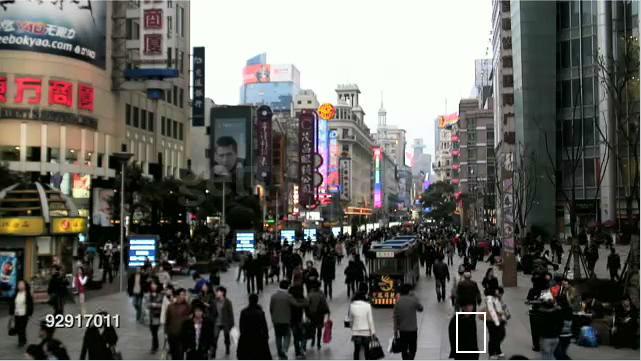} \\
\includegraphics[width=0.12\linewidth]{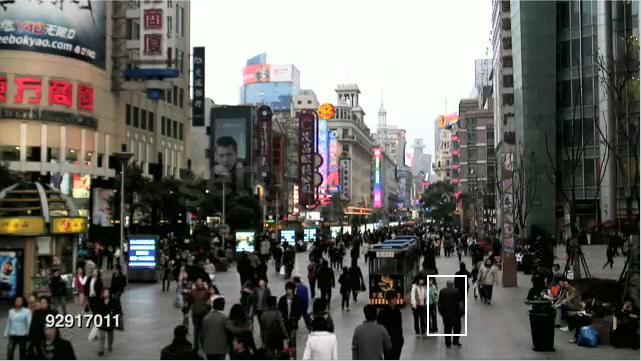} &
\includegraphics[width=0.12\linewidth]{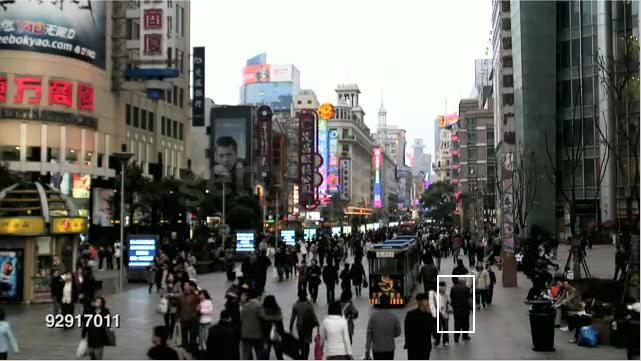} &
\includegraphics[width=0.12\linewidth]{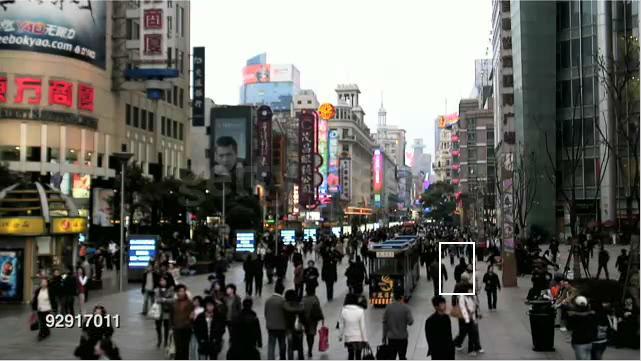} &
\includegraphics[width=0.12\linewidth]{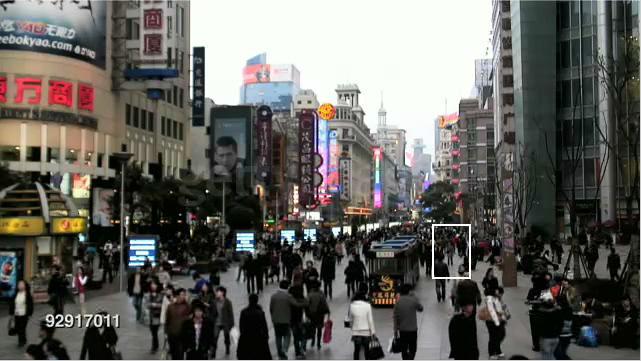} \\
\includegraphics[width=0.12\linewidth]{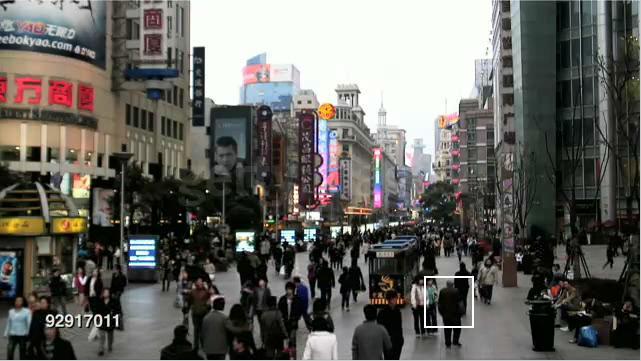} &
\includegraphics[width=0.12\linewidth]{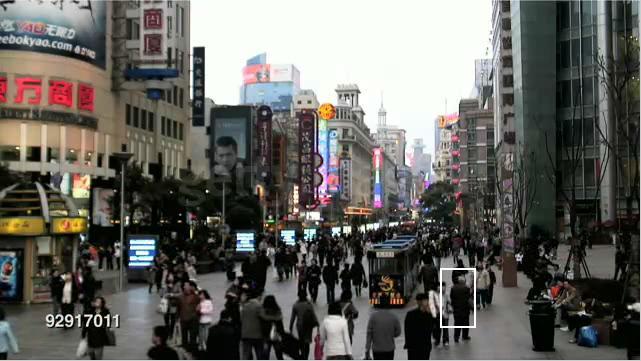} &
\includegraphics[width=0.12\linewidth]{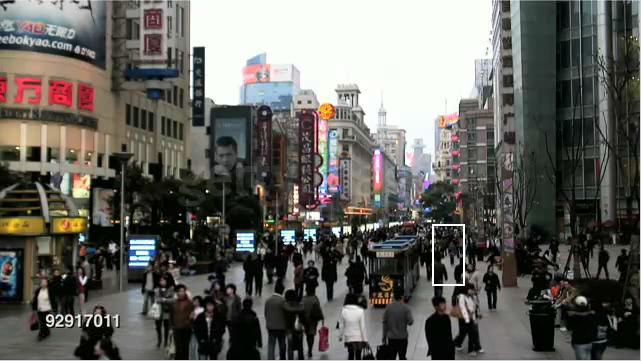} &
\includegraphics[width=0.12\linewidth]{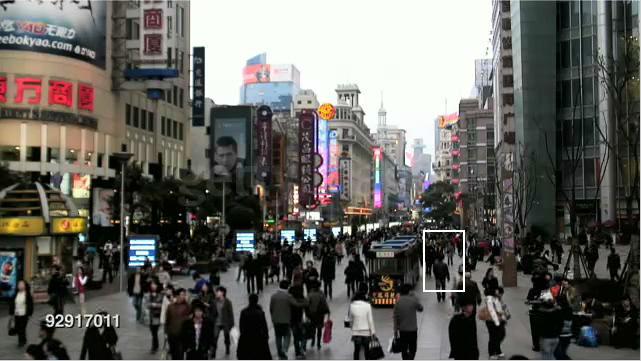} \\
\includegraphics[width=0.12\linewidth]{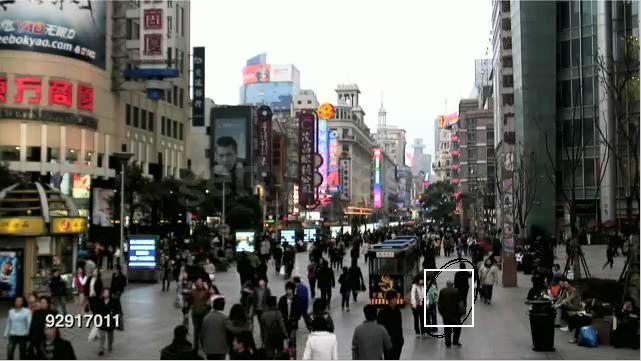} &
\includegraphics[width=0.12\linewidth]{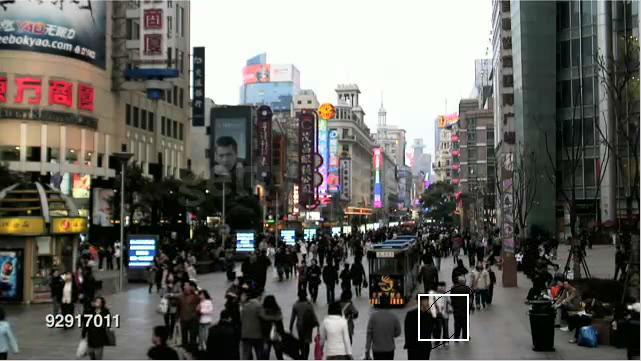} &
\includegraphics[width=0.12\linewidth]{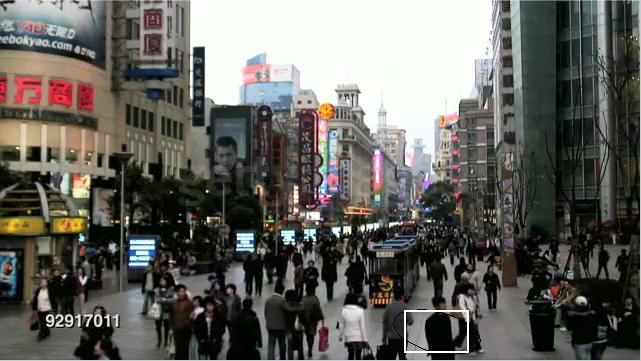} &
\includegraphics[width=0.12\linewidth]{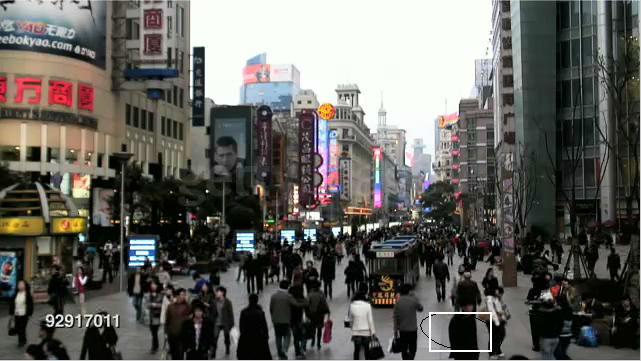} \\
\includegraphics[width=0.12\linewidth]{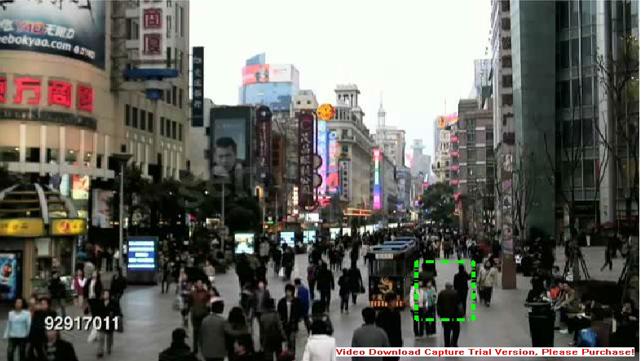} &
\includegraphics[width=0.12\linewidth]{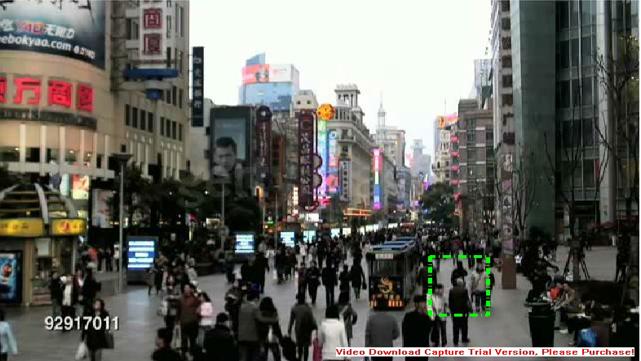} &
\includegraphics[width=0.12\linewidth]{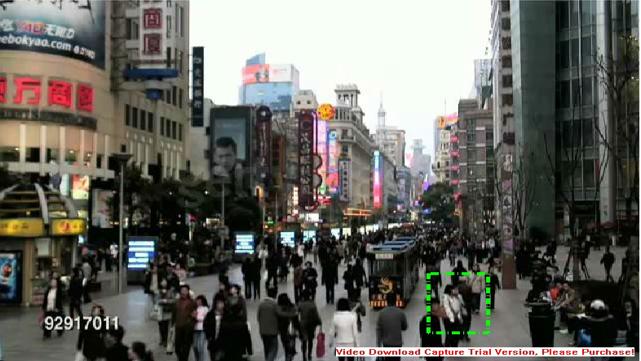} &
\includegraphics[width=0.12\linewidth]{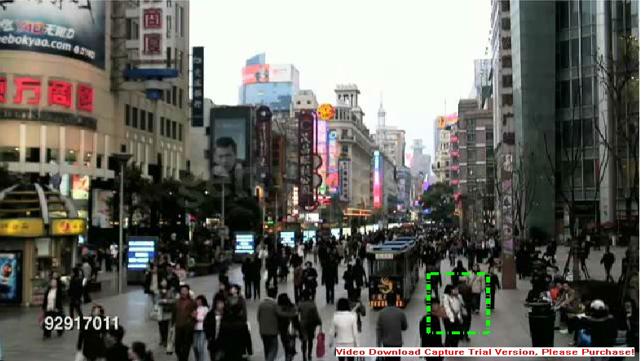} \\
\includegraphics[width=0.12\linewidth]{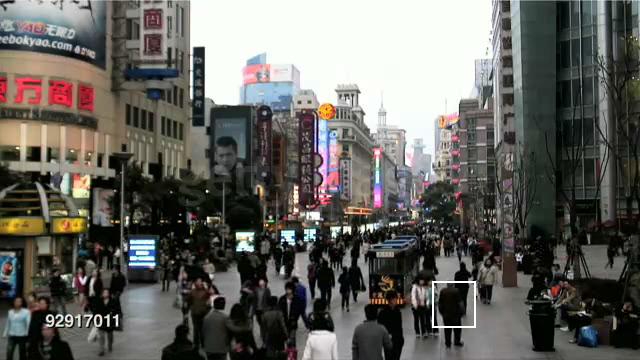} &
\includegraphics[width=0.12\linewidth]{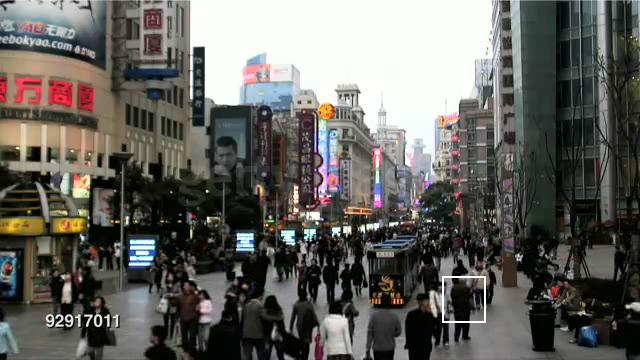} &
\includegraphics[width=0.12\linewidth]{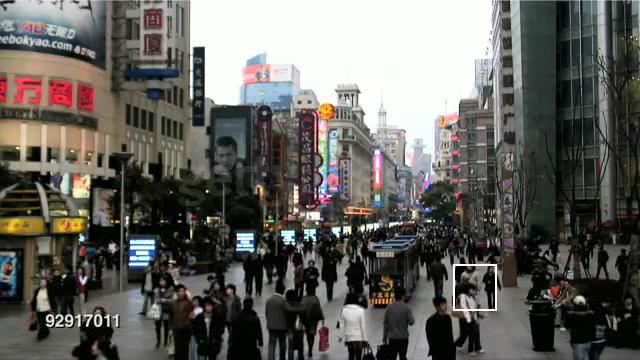} &
\includegraphics[width=0.12\linewidth]{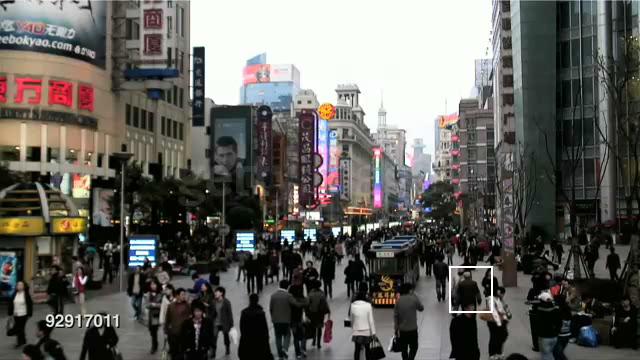} 
\end{tabular}}
\caption{Test sequences 4: Comparison of MS (1st row) \cite{comaniciu2003kernel}, EMS (2nd row) \cite{zivkovic2004like}, APF (3rd row) \cite{nummiaro2003adaptive}, MKF (4th row) \cite{zivkovic2009approximate}, CTM (5th row) \cite{trackunstr} and the proposed approach (Last row).}
\label{sequence4}
\end{figure*}
\subsection{Texture feature in mean shift framework}
The mean shift approach is based on the similarity measure of the probability density functions (PDF) of the  target model $\hat{O}$ i.e., template and candidate target model  $\hat{C}$ in the current frame. Here the similarity between the histograms is measured using the Bhattacharya coefficient \cite{Bhattacharya1}. The distance between the two PDFs is given as $\hat{d}(y)=\sqrt{1-\rho(y)}$.
and the Bhattacharyya coefficient $\rho(y)$ is given as below.
\begin{equation}
\rho [\hat{C}(y),\hat{O}]= \sum_{u=1}^k \sqrt{\hat{C}_u(y) \hat{O}_u}
\label{bhat_coff}
\end{equation} 
Here, $\hat{O}_u$ and $\hat{C}_u$ are the histograms with $k$ number of bins. These histograms denote the probability of number of feature vectors for a particular bin represented by $\hat{O}$. As these are the probabilities, their summation should be unity.
Equation (\ref{bhat_coff}) measures the similarity between the object model $O$ and the candidate model $C$ at location $y$. $\rho$ is the cosines of vectors given by $(\sqrt{O_1}, \sqrt{O_2}, \ldots ,\sqrt{O_k})^T $ and $ (\sqrt{C_1}, \sqrt{C_2}, \ldots , \sqrt{C_k})^T$.\\
This means the larger the $\rho$ (corresponds to or smaller the distance $\hat{d}(y)$), the higher will be the similarity or the good match between the target and object model. In order to find out the new target location, we try to maximize the Bhattacharyya coefficient. To maximize the (\ref{bhat_coff}) with respect to $y$ (target location)we employ the iterative mean shift algorithm \cite{comaniciu2003kernel}. Taking the first order Taylor series expansion of (\ref{bhat_coff}) the weights are obtained using the maximum likelihood estimation. This is given as the equation below.
\begin{equation}
w_i= \sum_{u=1}^k \hat{C}_u(y) \sqrt{\frac{\hat{O}_u}{\hat{C}_u(y_0)}}
\label{Weights}
\end{equation}
Here, $\hat{C}_u(y_0)$ is the initial estimate of the target and $\hat{C}_u(y)$ is the PDF at any general location $y$. The kernel density estimation of 
$\hat{C}_u(y)$ is defined in terms of the feature vector descriptor obtained using the texture curves as $\hat{C}_u(y)=\alpha \sum_{x_i \in S} K(||x_i||)^2 \delta[S(x_i)-u]$.
This equation describes the value of bin at $u$. Given the target, the texture curves are computed over dense 3 x 3 grid and $K(x)$ is the tracking kernel profile. The most common tracking kernel used is Epanechnikov kernel, which has been shown to be effective \cite{comaniciu2003kernel}. Now, the estimated weights are used in mean shift algorithm to compute the shift vector. Since $ \sum_{i=1}^n w_i(y_o)$ is strictly positive, mean shift vector can be written as $
M_h(y_0)=\left[\frac{\sum_{i=1}^n x_i w_i g\left(||\frac{y_0-x_i}{h}||\right)^2}{\sum_{i=1}^n w_i g\left(||\frac{y_0-x_i}{h}||\right)^2}\right] -y_0$.
Here, $g(x)$ is the derivative of Epanechnikov kernel $K$. Thus, the new target location is 
$\hat{y}=y_0+M_h(y_0)$.Thus in the current frame, the target position is computed as explained below.

 We calculate the weighted histogram of the target which depends on the tracking kernel. For the initial estimate of target, we calculate the histogram $C_u(y_0)$  and then compute the weights using similarity measures of the two histograms. The estimate of target in the current frame is computed using the mean shift vector. These steps are repeated until the algorithm converges.
\begin{figure*}[]
\centerline{%
\begin{tabular}{c@{\hspace{1pc}}c@{\hspace{1pc}}c@{\hspace{1pc}}c}
Frame 33 & Frame 42 & Frame 65 & Frame 88\\ \\
\includegraphics[width=0.12\linewidth]{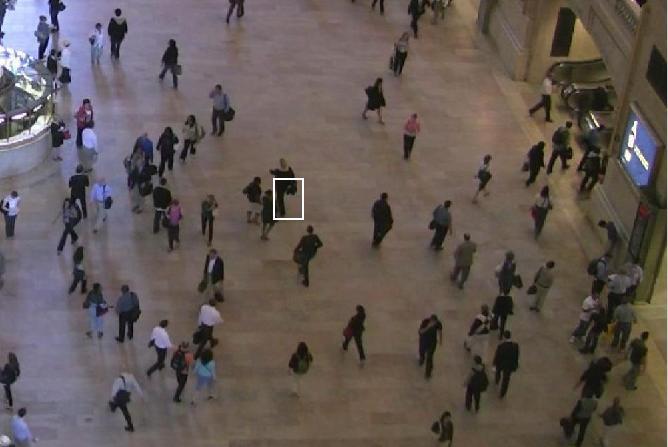} &
\includegraphics[width=0.12\linewidth]{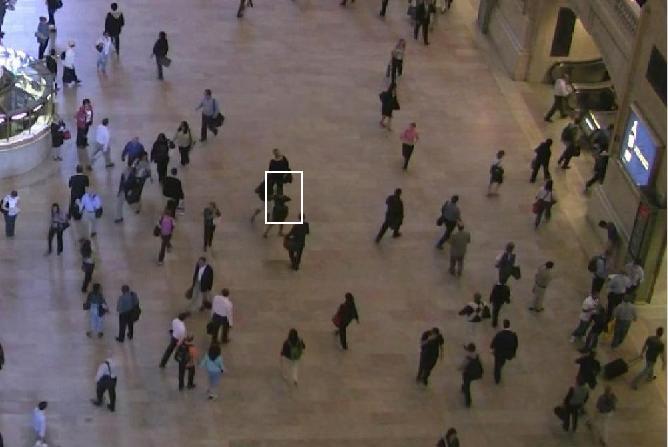} &
\includegraphics[width=0.12\linewidth]{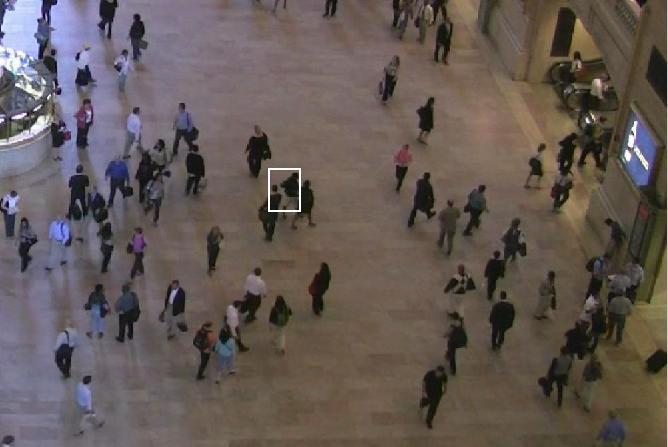} &
\includegraphics[width=0.12\linewidth]{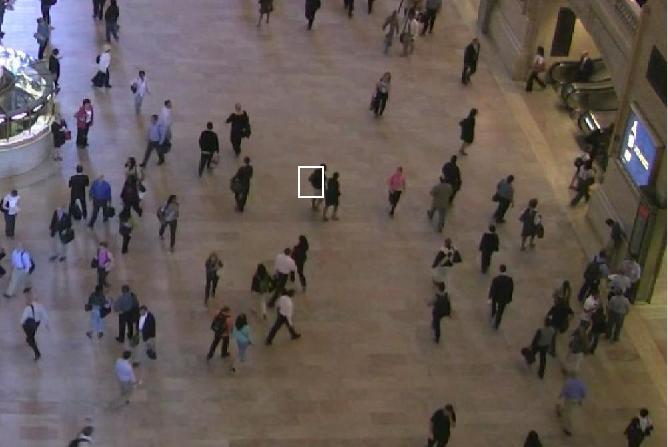} \\
\includegraphics[width=0.12\linewidth]{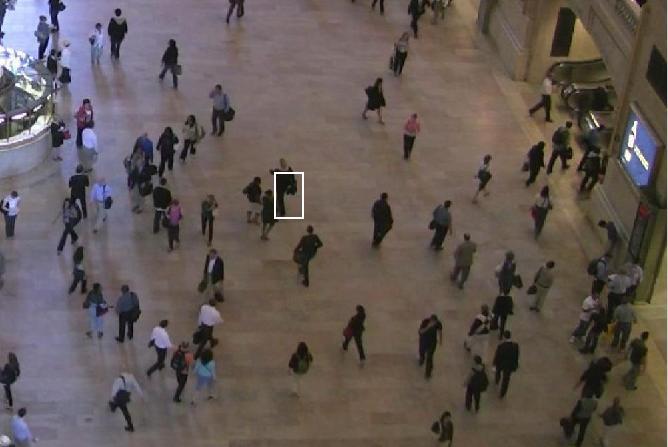} &
\includegraphics[width=0.12\linewidth]{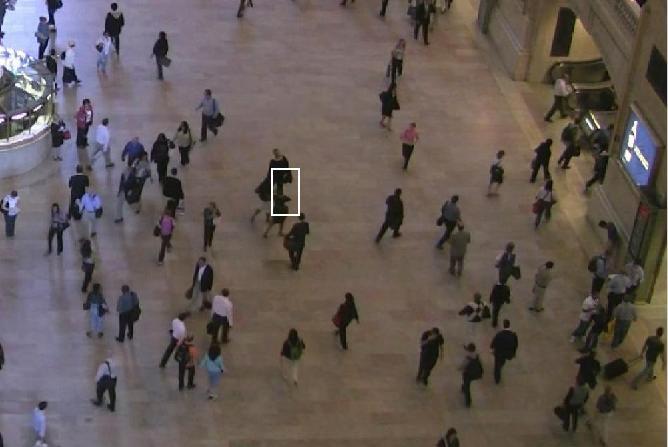} &
\includegraphics[width=0.12\linewidth]{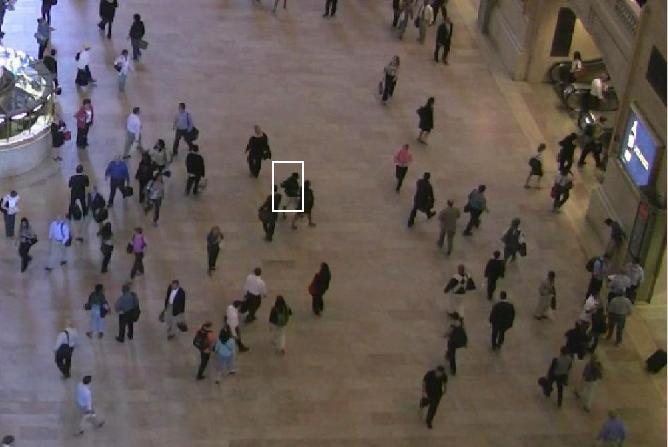} &
\includegraphics[width=0.12\linewidth]{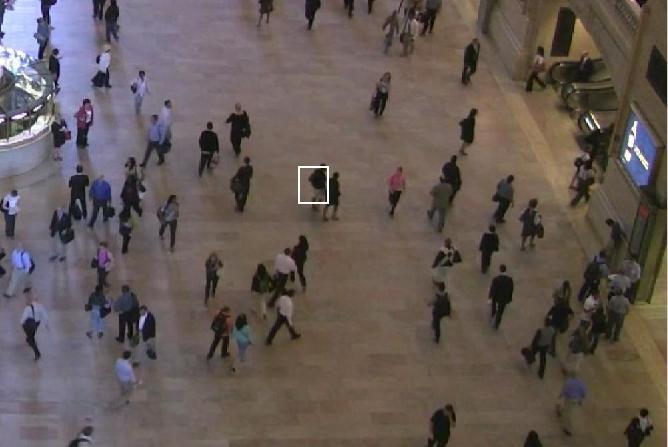} \\
\includegraphics[width=0.12\linewidth]{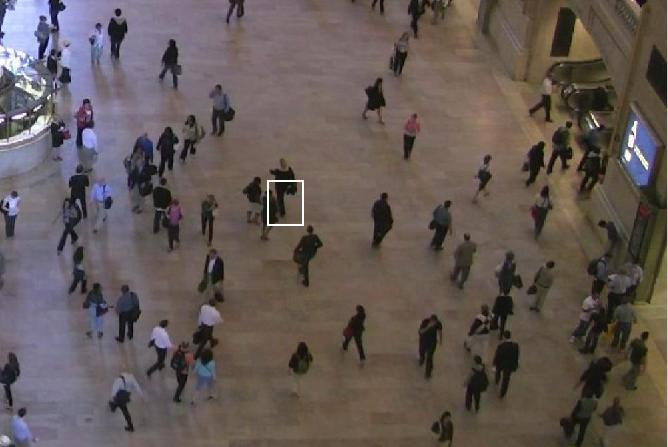} &
\includegraphics[width=0.12\linewidth]{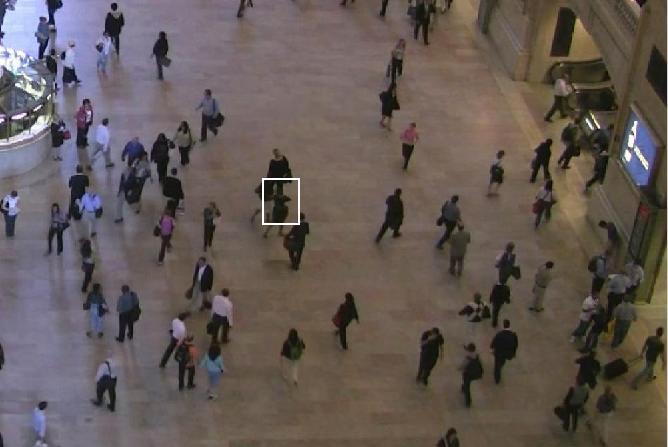} &
\includegraphics[width=0.12\linewidth]{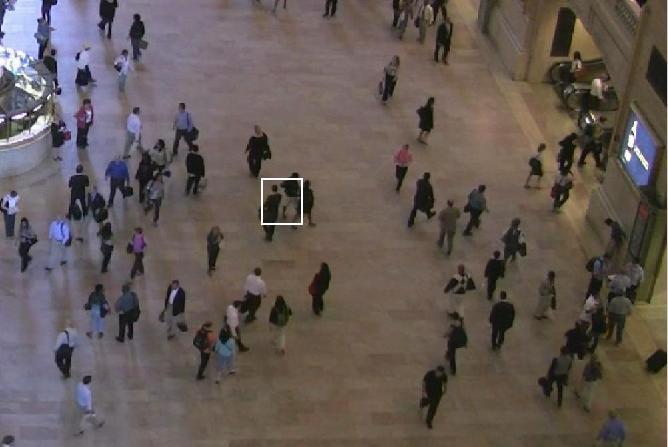} &
\includegraphics[width=0.12\linewidth]{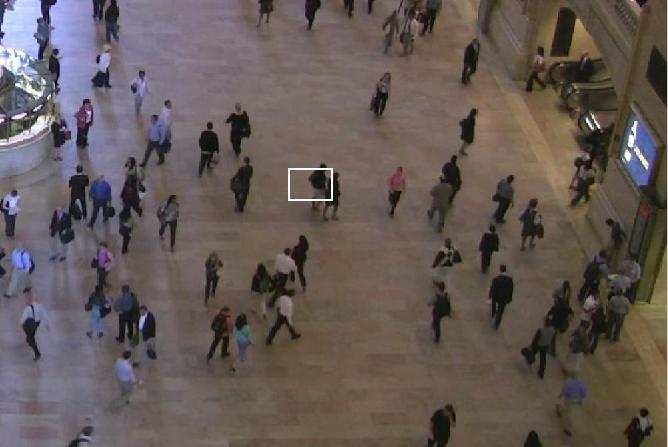} \\
\includegraphics[width=0.12\linewidth]{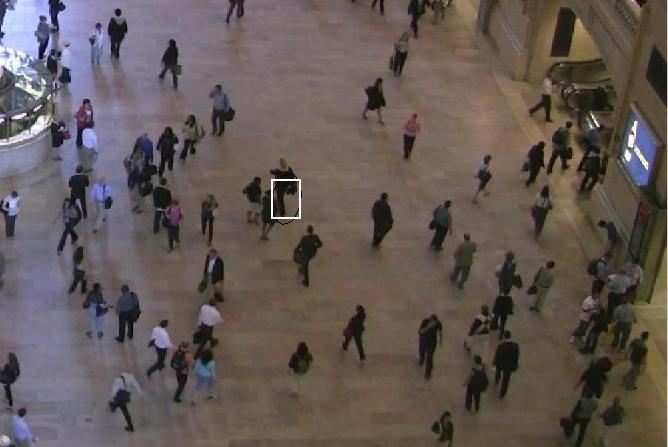} &
\includegraphics[width=0.12\linewidth]{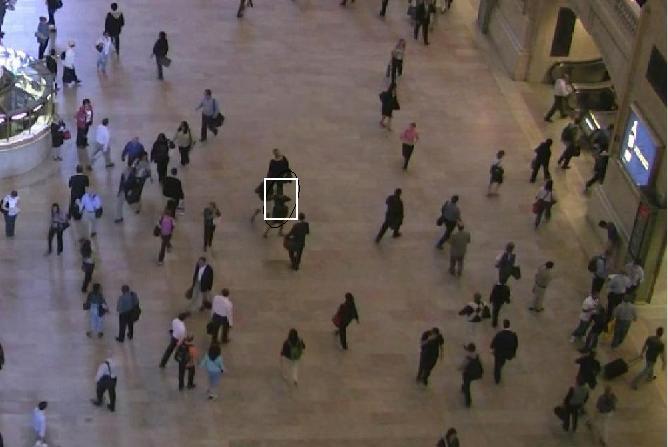} &
\includegraphics[width=0.12\linewidth]{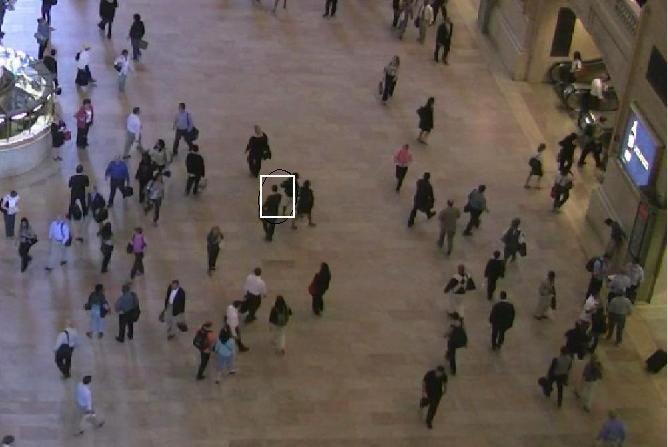} &
\includegraphics[width=0.12\linewidth]{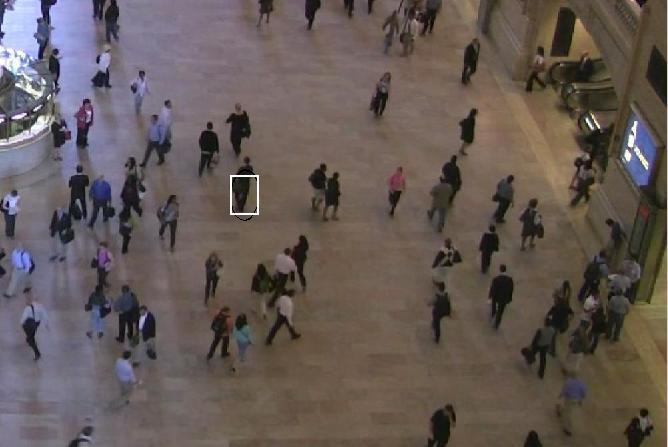} \\
\includegraphics[width=0.12\linewidth]{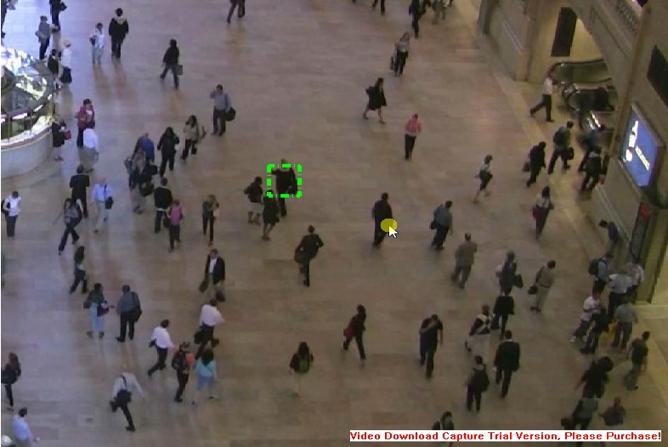} &
\includegraphics[width=0.12\linewidth]{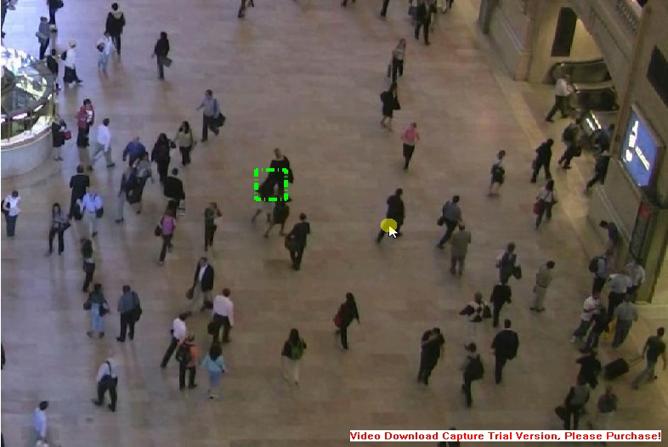} &
\includegraphics[width=0.12\linewidth]{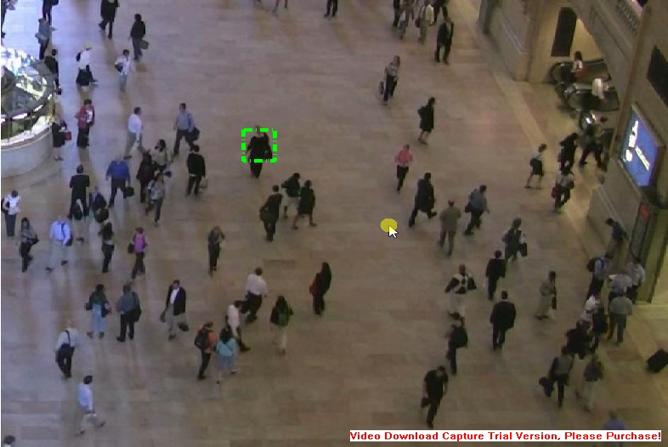} &
\includegraphics[width=0.12\linewidth]{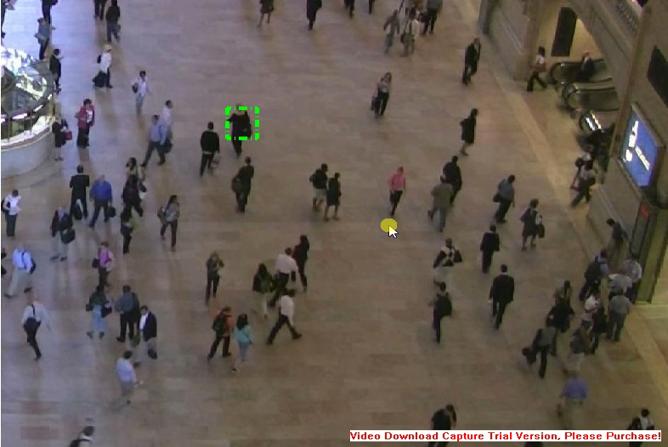} \\
\includegraphics[width=0.12\linewidth]{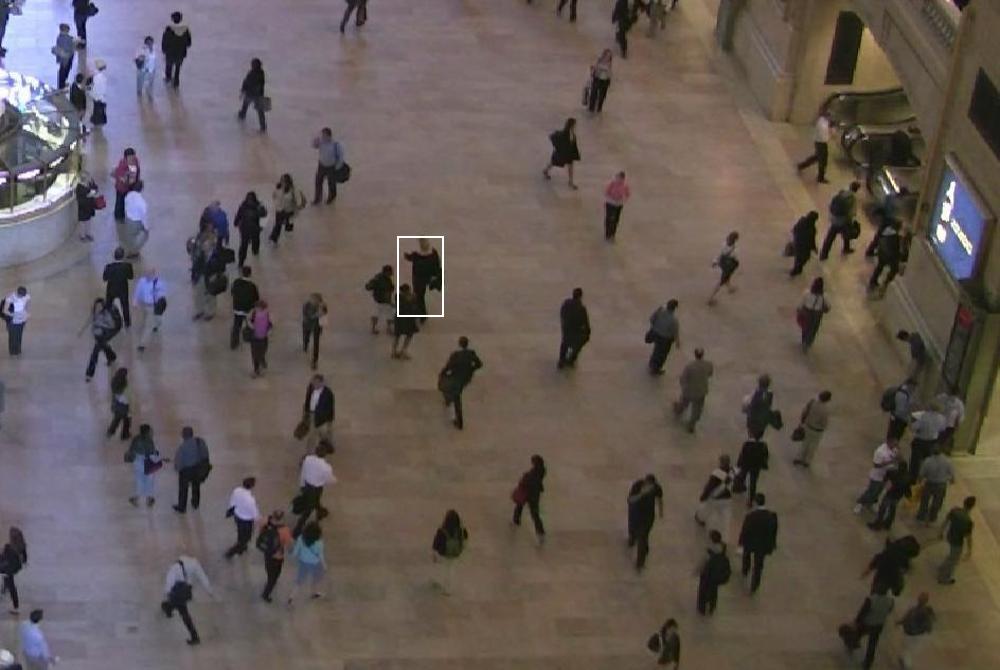} &
\includegraphics[width=0.12\linewidth]{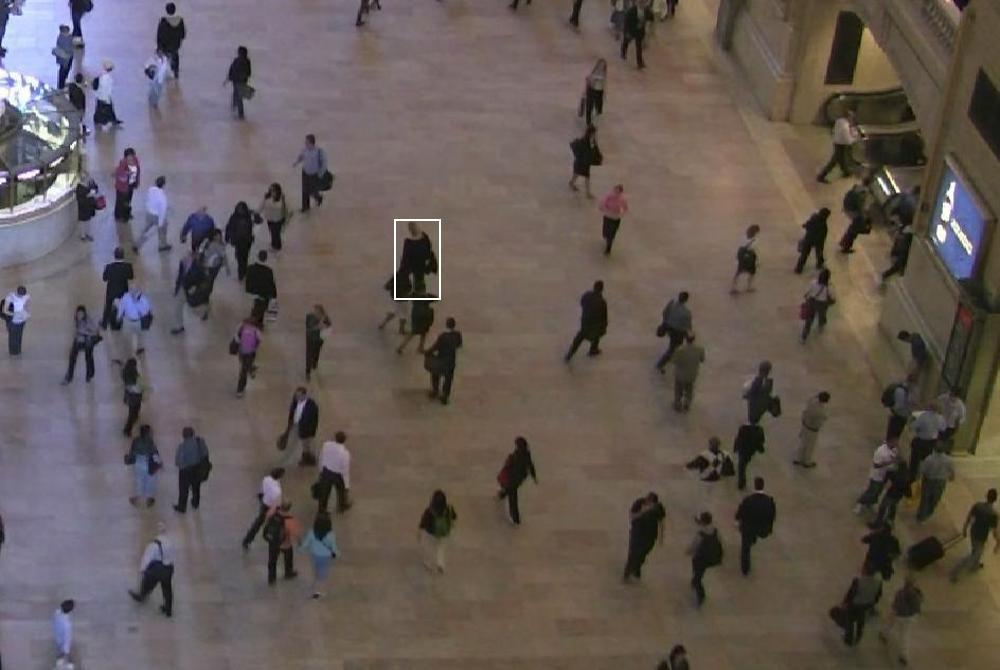} &
\includegraphics[width=0.12\linewidth]{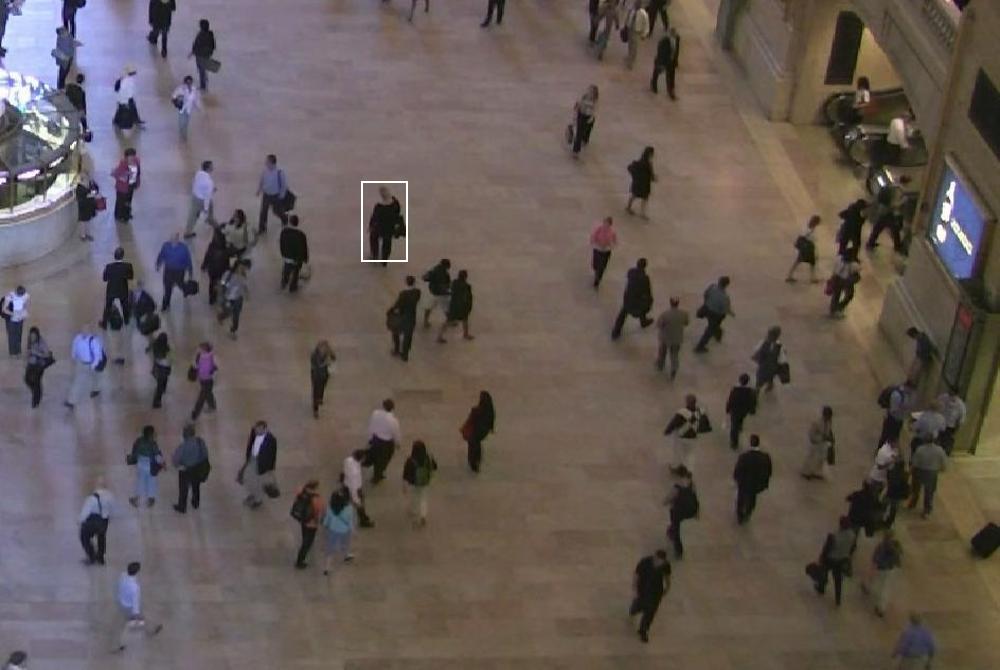} &
\includegraphics[width=0.12\linewidth]{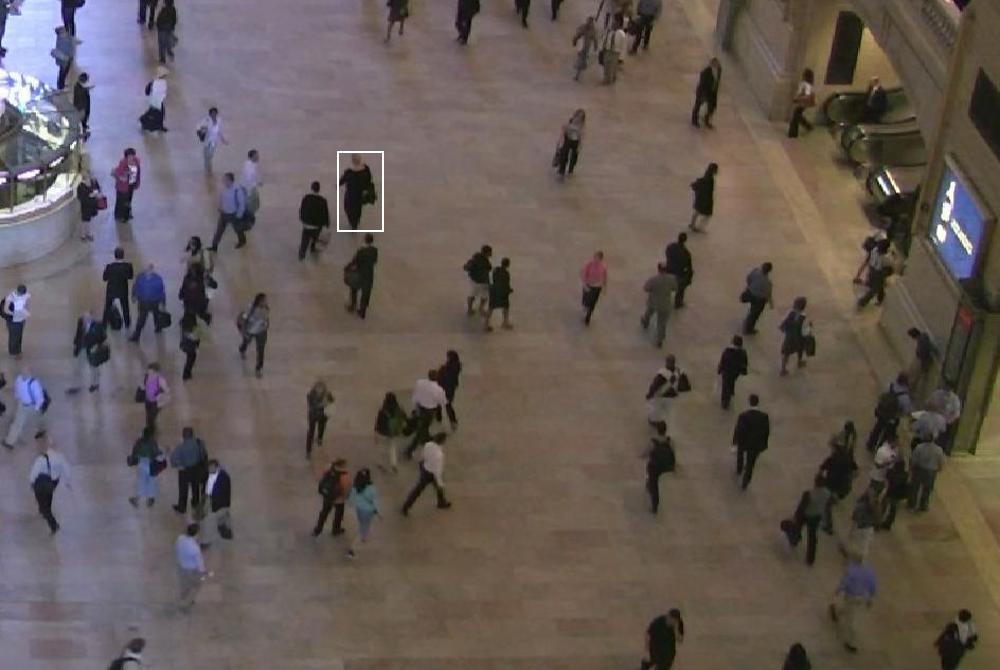}
\end{tabular}}
\caption{Test sequences 1: Comparison of MS (1st row) \cite{comaniciu2003kernel}, EMS (2nd row) \cite{zivkovic2004like}, APF (3rd row) \cite{nummiaro2003adaptive}, MKF (4th row) \cite{zivkovic2009approximate}, CTM (5th row) \cite{trackunstr} and the proposed approach (Last row).}
\label{sequence1}
\end{figure*}

\section{Results and Discussion}
Data used in this work comprise of videos of shopping malls and busy streets and all the videos show a number of  interactions between the agents. These videos are of unstructured crowd scenes where the agent moves in complex pattern as well as cluttered by other agents. Data used in our work are obtained from \cite{rodriguez2011data, ali2007lagrangian} and the performance of our algorithm has been evaluated on this data.  
Results of our algorithm is compared with the state-of-the-art mean shift tracker (MS) based on RGB color values \cite{comaniciu2003kernel}, extended mean shift tracker (EMS) \cite{zivkovic2004like}, multiple Kalman filters tracker (MKF) \cite{zivkovic2009approximate}, adaptive color based particle filter (APF) \cite{nummiaro2003adaptive}, and correlated topic model tracker (CTM) \cite{trackunstr}. For color based trackers MS, EMS, MKF, APF all the three color channels were quantized into $16$ indexes and a histogram is generated having $16 \times 16 \times 16$ number of bins. For the CTM Tracker we are using the software provided by the author. In our algorithm, we use OTC descriptor to describe the target and a histogram is generated using the visual vocabulary encoding. Here the vocabulary is obtained by using K-means clustering and the size of vocabulary varies from 100 to 500 words, typically 100 word size vocabulary has been chosen for further processing.

We'll discuss the results of the proposed approach on videos obtained from the different scenes as shown in Fig. \ref{target} and Fig. (\ref{sequence1}-\ref{sequence4}) demonstrate the merits and demerits of the proposed approach. 

Tracking results of first test sequence are illustrated in Fig. \ref{sequence1}. These images show the sequence of a busy shopping mall. In this video, target object is far from the camera and is moving in a particular direction. Here, first frame (Frame 33) describes the time frame when the target object starts interacting with the other agents and from the first column it can be very easily observed that every algorithm including the proposed approach work well upto Frame 33. When the interaction is over (Frame 42) only the proposed approach and the tracker introduced in \cite{trackunstr} are able to track the target object as illustrated in second column. It is because of the fact that the tracked object is interacted with another agent having the same color that of the target and all the color based trackers failed whereas the directional sensitivity of the proposed approach forced it to track the target object moving in a particular direction. Most of the trackers start tracking the object which interacted with the target object as shown in the third and fourth column (Frame 65, 88) which is not desired for video based surveillance applications.

Test sequence as shown in Fig. \ref{sequence2} show the images of the airport terminal lobby and in this case, the target is far from the camera and interacts with a number of other agents. As the color of the target object is different from the interacting agents so all the trackers expect APF and MKF perform well upto the third column (Frame 43) including the proposed approach. In the last column (Frame 63), it can be observed that all the trackers lost the target object including the CTM tracker \cite{trackunstr}. However, the proposed approach successfully tracked the target throughout the video because the proposed texture descriptor can easily differentiate between the texture and texture-less areas. Further, codebook encoding enables us to gather all texture-less and textured regions together.

In the third scenario, shown in Fig. \ref{sequence3}, an object is tracked in highly cluttered environment where the target is passed across by a number of agents at different instants of time. In this particular example, our interest is to track a person wearing blue shirt and black coat who then interacts with the other persons wearing black clothes. Except the proposed algorithm (last row) and APF (3rd row) tracker, all the trackers lost the target - some lost mid way and some towards the end. Both the EMS (2nd row) and the MKF (4th row) tracker lost the target in mid way but returned back to the target object at the end because of the color sensitivity on the target. The CTM tracker results in poor tracking as shown in the fifth row of  Fig. \ref{sequence3} and starts tracking some other agent. However, it can be observed that the proposed approach tracks the object throughout the video effectively. 

Fig. \ref{sequence4} shows a sequence of a busy street. In this sequence an object is tracked which is completely occluded by the other agents of totally different colors. It can be clearly observed from third column (Frame 112) of the Fig. \ref{sequence4} that all the trackers fail after the interaction, except the proposed approach which tries to maintain the directionality and texture features of the tracked object. 

As all the videos depict the real world surveillance settings, the results obtained using our approach is far better compared to the state-of-the-art tracking approaches. Further, all the videos we have shown are surveillance based and hence captured using a static camera. We are trying to improve our method to even take of cameras which undergo motion. 

\section{Conclusion}
We have proposed a framework to track an individual in less dense but highly unstructured crowd scenes by incorporating OTC descriptor in mean shift tracking framework. In contrast to other available methods for tracking, our method performs well even when there is interaction among the agents and when the agent follows a complex motion pattern. A number of experiments have been performed to compare the proposed approach with the number of other trackers over a dataset of less dense unstructured crowd scenes. We found that using the rotational sensitive texture information in tracking framework works effectively. We would like to consider more dense crowd scenes and modify the proposed approach to track objects in such scenes. We also plan to extend the approach to crowd scenes captured using hand-held cameras. We have shown that the proposed approach is an efficient step towards achieving robust tracking of objects in unstructured crowd scenes.



%

\bibliographystyle{./IEEEtran}
\bibliography{pub_tex} 

\end{document}